\documentclass[twoside,11pt,numbers]{article}
\usepackage[utf8]{inputenc}
\usepackage{longtable} 
\usepackage{textcomp} 
\usepackage{ragged2e} 
\usepackage[a4paper, margin=1in]{geometry} 
\usepackage{array}
\usepackage{blindtext}
\usepackage{longtable} 
\usepackage{comment}   
\usepackage{algorithm}
\usepackage{algpseudocode}
\usepackage{multirow}
\usepackage{booktabs}       
\usepackage{amsmath} 

\usepackage{tikz}
\allowdisplaybreaks
\usetikzlibrary{positioning, arrows.meta, fit, backgrounds}

\makeatletter
\def\l@algorithm{\@dottedtocline{1}{\z@}{}} 
\def\plist@algorithm{Algorithm\space}
\makeatother

\let\oldbibliographystyle\bibliographystyle
\renewcommand{\bibliographystyle}[1]{}

\usepackage{dmlr2e}

\renewcommand{\citep}{\citeyearpar}

\usepackage{lastpage}

\ShortHeadings{Rethinking Causal Discovery}{T. Brogueira and M. Figueiredo}
\firstpageno{1}

\begin{document}

\title{Rethinking Bivariate Causal Discovery Through the Lens of Exchangeability}

\author{\name Tiago Brogueira \email tiago.brogueira@tecnico.ulisboa.pt \\
       \addr Instituto de Telecomunicações\\
       Instituto Superior Técnico\\
       Lisboa, Portugal
       \AND
       \name Mário Figueiredo \email mario.figueiredo@tecnico.ulisboa.pt \\
       \addr Instituto de Telecomunicações\\
       Instituto Superior Técnico\\
       Lisboa, Portugal}

\maketitle

\begin{abstract}
{Causal discovery methods have traditionally been developed under two different modeling assumptions: \textit{independent and identically distributed} (i.i.d.) data and time series data. In this paper, we focus on the i.i.d. setting, arguing that it should be reframed in terms of exchangeability, a strictly more general symmetry principle. For that goal, we propose an exchangeable hierarchical model that builds upon the recent \textit{Causal de Finetti theorem}. Using this model, we show that both the uncertainty regarding the causal mechanism and the uncertainty in the distribution of latent variables are better captured under the broader assumption of exchangeability. In fact, we argue that this is most often the case with real data, as supported by an in-depth analysis of the T\"ubingen dataset. Exploiting this insight, we introduce a novel synthetic dataset\footnote{All our code is open-source and available at \href{https://github.com/tiagobrogueira/Thesis-Causal-Discovery-In-Exchangeable-Data/tree/main}{https://github.com/tiagobrogueira/Thesis-Causal-Discovery-In-Exchangeable-Data/tree/main}} that mimics the generation process induced by the proposed exchangeable hierarchical model. We show that our exchangeable synthetic dataset mirrors the statistical and causal structure of the T\"ubingen dataset more closely than other i.i.d. synthetic datasets. Furthermore, we introduce SynthNN, a neural‐network–based causal‐discovery method trained exclusively on the proposed synthetic dataset. The fact that SynthNN performs competitively with other state‐of‐the‐art methods on the real‐world T\"ubingen dataset provides strong evidence for the realism of the underlying exchangeable generative model.}

\end{abstract}

\section{Introduction}

The aim of scientific research is often to find causal relationships between variables of interest. This stems either from a desire to intervene in the system under study (and not merely be able to make accurate statements over the data we observe) or from a goal of uncovering causal mechanisms underlying the observations, seeking a deeper understanding of the underlying phenomena \citep{pearl2009causality}. Traditionally, these relationships are uncovered by performing experiments (known in this context as \textit{interventions}). However, this can often be impossible, impractical, or unethical \citep{glymour2019review}. Furthermore, if not all intervening variables are controlled for, this can lead to erroneous results (e.g., Simpson's paradox \citep{ameringer2009simpson}). In these situations, the need arises to learn causal relationships from observational data alone. This growing field, which uses a purely data-driven approach to learn causal structure (represented using a graph), is called \textit{causal discovery}.

Causal discovery methods are typically divided into two main categories: those assuming independent and identically distributed (i.i.d.) data and those designed for time series data \citep{timeseriesvsiid}, which will not be addressed in this paper\footnote{From this point onward, we will often refer simply to causal discovery, but it should be interpreted as referring to the family of causal discovery approaches for data traditionally assumed to be i.i.d.} \citep{timeseriesvsiid2}.  An important result in this field is that, when working with purely observational i.i.d. data and no further assumptions, one can only identify the causal structure up to the so-called \textit{Markov equivalence class} (MEC). This is defined as the set of causal graphs that satisfy the same conditional independence properties \citep{spirtes2001causation}. In particular, in the case of a pair of dependent variables, $X$ and $Y$, both causal directions ($X \rightarrow Y$, $Y \rightarrow X$) belong to the same MEC, as there are no conditional independence properties to be satisfied.  Furthermore, under the i.i.d. observations assumption, everything not explained by the causal mechanism is considered as noise, often formalized as exogenous variables \citep{zanga2022survey}. \label{MEC}

In general, identifying one of the elements of the MEC requires \textit{interventions} \citep{pearl2009causality}. Formally, interventions correspond to modifications of the base structural causal model by altering or replacing the causal mechanisms (i.e., the incoming functions) for one or more chosen variables. Each such intervention defines a new “environment” whose data are still drawn i.i.d. from the intervened model\label{interventions}. 

Recently, the concept of exchangeability was brought into the field of causal discovery by Guo et al. \citep{cdf}. The main motivation of those authors is the fact that, by considering the data to be exchangeable, this allows more complex dependencies between the data points to be explored, even in the bivariate case. By treating bivariate data as an exchangeable sequence rather than strictly i.i.d., one can analyze relationships across multiple observations. This broader perspective unlocks cross-sample conditional independencies that break the observational symmetry, enabling the identification of the causal direction. Those authors propose a ``Causal de Finetti" theorem, which demonstrates that exchangeable data containing specific cross-sample conditional independencies naturally represent independent causal mechanisms. 

In this paper, we argue that the advantage of considering the data to be exchangeable goes beyond the original motivation of Guo et al. \citep{cdf}. Specifically, we show that it is also a more realistic model of real-world cause-effect pairs. Moreover, we enrich the exchangeable model of Guo et al. with a hierarchical structure, improving the flexibility of the exchangeability-based approach to bivariate causal discovery and of the Causal de Finetti theorem.

The rest of the paper is organized as follows. In Section 2, we begin by laying out the problem and reviewing previous work. In Section 3, we define and argue for the new hierarchical exchangeable model that better represents real-world causal mechanisms. Section 4 describes a new synthetic dataset generated in accordance with the proposed hierarchical model and compares it to other datasets. Finally, Section 5 introduces a neural network trained exclusively on this synthetic dataset, which serves as a new valid causal discovery method, and, most importantly, as further validation of the proposed model and its assumptions and rationale.

\section{Causal Discovery, Exchangeability, and Their Intersection}
\subsection{Causal Discovery}
Causal discovery is the problem of determining the causal relationships among a collection of dependent random variables of interest, say $X_{(1)},...,X_{(d)}$, usually described using a \textit{structural causal model} (SCM) \citep{pearl2009causality}. Typically, an SCM includes a \textit{directed acyclic graph} (DAG) that specifies, for each $X_{(i)}$, what is the collection of other variables that are directly causally related to $X_{(i)}$, called the parents of $X_{(i)}$ and denoted $Pa(X_{(i)})$.  Functional causal models (FCM) are a particular type of SCM where the dependency of each $X_{(i)}$ on its parents takes the form
\begin{equation}
    X_{(i)} = f_i (Pa(X_{(i)}),\epsilon_i),
\end{equation}
where $\epsilon_1,...,\epsilon_d$ is a collection of so-called \textit{exogenous variables}, which are considered to be mutually independent \citep{pearl2009causality, icm, zanga2022survey}

Causal discovery from purely observational data is an important, diverse, and large research field, which would be impossible to comprehensively review here (see \citep{zanga2022survey}, for a recent survey and pointers to a vast literature). In this paper, we focus on what is arguably the quintessential causal discovery problem, called \textit{bivariate causal discovery} or the \textit{cause-effect problem}  \citep{tuebingenresults,tuebingen}. Given two dependent variables $X$ and $Y$ (in practice, samples thereof), the famous Reichenbach \textit{common cause assumption} states that "there is no dependency without causation;" that is, either there is a third variable that is a \textit{common cause} for $X$ and $Y$, called a confounder, $X$ causes $Y$, or $Y$ causes $X$. If there are reasons to assume \textit{causal sufficiency} (\textit{i.e.}, absence of an unobserved \textit{confounder}) and no feedback \citep{tuebingen}, the problem becomes simply to select between the two possible causal directions: $X$ causes $Y$ or $Y$ causes $X$.

As mentioned in Section \ref{MEC}, in the bivariate case, it is impossible to distinguish between the two possible causal directions without interventions or additional assumptions. In the purely observation scenario, different methods are based on different assumptions, which typically take the form of some property that holds in the true causal direction but not in the reverse direction. Some methods exploit the independence between different variables, for example, independence between the noise exogenous variable and the cause \citep{peters2010identifying} or independence between noise and function mechanism \citep{janzing2012information}. Another family of assumptions is based on Occam's razor, restricting the set of accepted causal mechanisms and choosing the causal direction that has a better fit. For example, Blöbaum et al. \citep{reci} restrict the functions to be linear, Goudet et al. \citep{cgnn} opt for a neural network with one hidden layer, while Dhir et al. use Bayesian model selection with priors on the functions to be more flexible \citep{bayes}. Lastly, some methods are based on analysing the complexity of the model in each direction, assuming the model in the true direction has a lower Kolmogorov complexity; however, since Kolmogorov complexity is not computable, it is necessary to resort to different approximations of this measure \citep{slope,bcqd}.

\label{Tübingen_intro}
Regrettably, the causal discovery research area is critically lacking in extensive real‐world datasets\footnote{Here, “extensive” denotes the presence of many distinct examples. Although numerous multivariate datasets exist, each comprises only a single causal graph, which precludes the use of evaluation metrics such as accuracy.}, with the Tübingen dataset being the only one widely accepted in the literature \citep{tuebingen}. It contains a total of 108 cause-effect pairs with known ground truth based on expert knowledge. These pairs were gathered from 37 different domains; some examples are the altitude and temperature of cities, the horsepower and fuel consumption of cars, and the age and height of different people. Given the lack of good real-world benchmarks, it is common to also test new methods on synthetic datasets. For the bivariate problem in particular, four commonly used datasets are CE-Cha, CE-Net, CE-Gauss, and CE-Multi \cite{tuebingenresults}. These datasets (except CE-Cha) were designed under certain assumptions (such as additive or multiplicative noise) and thus are usually used to test whether certain methods can detect the true causal direction under those assumptions, not to assess the more general performance of the methods. 

\subsection{Exchangeability}
The concept of exchangeability, deeply tied to the foundational representation theorem by de Finetti, corresponds to a probabilistic notion of (permutation) symmetry \citep{definetti}. Intuitively, it establishes the conditions under which the order in which a sequence of observations is collected is irrelevant.

\textbf{Definition\label{exchangeref}:} \textit{An exchangeable sequence of random variables is a finite or infinite sequence \(X_1, X_2, X_3, \ldots\), such that for any finite permutation \(\pi:\{1, \ldots, N\} \rightarrow \{1, \ldots, N\} \) of the position indices, the joint distribution of the permuted sequences is the same as that of the original:}
\begin{equation}
    P(X_{\pi(1)}, \ldots, X_{\pi(N)}) = P(X_1, \ldots, X_N).
\end{equation}
Additionally, a sequence is said to be \textit{partially exchangeable} if there exists a partition of the indices such that permutations within each partition are exchangeable.

Naturally, all independent and identically distributed (i.i.d.) sequences are exchangeable, but, of course, the converse is not true. Also, as the name implies, time series are obviously not exchangeable sequences since the order of the samples plays a central role in modeling this type of data. The following example aims to illustrate the concept of exchangeability. 

\vspace{0.1cm}

\begin{description}\label{exchangeable_example}
    \item[Exchangeable urn sequence example:] Consider two urns (denoted as $A$ and $B$) with different ratios of black and white balls to the total number of balls (denoted $\theta_A$ and $\theta_B$). At the beginning of the experiment, a fair coin is flipped to select one of the urns, but which urn was chosen is not revealed to the observer. Subsequently, balls are randomly drawn with replacement from the selected urn. Let $X_i \in \{ 0, \, 1\}$ denote the $i$-th sample, with 0 and 1 denoting white and black balls, respectively. The sequence thus generated is exchangeable; for example, 
    \[
    P(X_1 = 1, X_2=0) = \theta_A (1-\theta_A) \tfrac{1}{2} + \theta_B (1-\theta_B)\tfrac{1}{2}  = P(X_1 = 0, X_2=1).
    \]
    However, the samples are not independent; for example, since $P(X_t = 1) = \tfrac{1}{2} \theta_A + \tfrac{1}{2} \theta_B$, for any $t$,
     \[ 
     P(X_1 = 1)\, P(X_2 = 0) =  (\tfrac{1}{2} \theta_A + \tfrac{1}{2} \theta_B)(1 - \tfrac{1}{2} \theta_A - \tfrac{1}{2} \theta_B) \neq  P(X_1 = 1, X_2=0),
     \]
    thus the sequence is not i.i.d. (except, of course, in the trivial case where $\theta_A = \theta_B$).
\end{description}

\vspace{0.1cm}
The most important result regarding exchangeability is given by the famous \textit{de Finetti representation theorem} \citep{definetti}, stated next.
\begin{description}
    \item[De Finetti theorem:] Let \((X_n)_{n \in \mathbb{N}}\) be an infinite sequence of binary random variables. The sequence is exchangeable if and only if there exists a random variable \(\theta \in \Theta\), with probability measure \(\mu\), such that \(X_1, X_2, \dots\) are conditionally i.i.d. given \(\theta\), that is, for any given any sequence \((x_1, \dots, x_N) \in \{0, 1\}^N\), 
\begin{equation} \label{definettiref}
    P(X_1, \ldots, X_N) = \int_\Theta \prod_{n=1}^N P(X_n | \theta) \, d\mu(\theta).
\end{equation} 
\end{description}
In Equation \eqref{definettiref}, $d\mu(\theta)$ can be replaced by $p(\theta)\, d\theta$, if $\mu$ is absolutely continuous with respect to the Lebesgue measure on $\Theta$, \textit{i.e.}, if there is a probability density function $p(\theta))$. In the case where $\Theta$ is a discrete set, Equation \eqref{definettiref} takes the form 
\begin{equation} \label{definettiref_discrete}
     P(X_1, \ldots, X_N) = \sum_{\theta\in \Theta} \prod_{n=1}^N P(X_n | \theta) \, p(\theta),
\end{equation} 
where, in this case, $p(\theta)$ is a probability mass function. The theorem has been extended in several ways, including to non-binary variables \citep{bayesiantheory}. In essence, this theorem states that any sequence of exchangeable variables can be seen as a mixture of i.i.d. sequences, each of which is conditioned on the parameter $\theta$.

\vspace{0.1cm}

\begin{description}\label{exchangeable_example}
    \item[Exchangeable urn sequence example (continued):] The example above illustrates the de Finetti theorem with $\Theta =\{\theta_A,\, \theta_B\} \subset [0,1]$, $p (\theta_A) = p(\theta_B) = 1/2$, and $P(X_n | \theta ) = \theta^{X_n}\, (1-\theta)^{1-X_n}$.
\end{description}

\vspace{0.1cm}

The existence of the latent (unobserved) parameter $\theta$ mediates the dependence between different variables in an exchangeable sequence. In the urn experiment, for example, each draw provides information about this underlying parameter, which, in turn, informs our expectations about future draws. In Bayesian theory, $\theta$ is treated as a latent variable, and the distribution $p_{\Theta}(\theta)$, which emerges naturally from de Finetti's theorem, acts as the Bayesian prior \citep{bayesiantheory}. Thus, the requirement for a prior in Bayesian inference does not stem from mathematical convenience or a lack of knowledge, but from the fundamental modeling of $\theta$ as a latent variable \citep{bayesianproof}. Ultimately, this demonstrates that it is our judgment of exchangeability, not any metaphysical belief about a ``true'' model, that underpins the standard use of Bayesian modeling involving i.i.d. observations conditioned on an unknown latent variable.

\subsection{Hierachical Models}
\label{sec:hier}
In this subsection, we enrich the class of models described above with additional latent variables, yielding so-called \textit{hierarchical models} \citep{bayesianproof}. The rationale for the introduction of these hierarchical models is as follows: sometimes, a complicated distribution $P(X|\theta)$ can be conveniently represented as the marginal with respect to some latent variable $\eta$, $P(X|\theta) = \int P(X|\eta,\theta) p(\eta|\theta)\, d\eta$; mixture models are a classical example. 

The complete family of models is given by the following set of instances of the de Finetti representation, which we will comment on in detail below:
\begin{align}
    \text{Model 1:} \;\; P(X_1, \ldots, X_N | \theta) &=   \prod_{n=1}^N P(X_i | \theta), \label{eq:model1}\\
    \text{Model 2:} \;\; P(X_1, \ldots, X_N) &= \int_\Theta \prod_{n=1}^N P(X_n | \theta) \, p(\theta)\, d\theta, \label{eq:model2} \\
    \text{Model 3:} \;\; P(X_1, \ldots, X_N | \theta) &= \prod_{n=1}^N \int_H P(X_n | \eta_n) \, p(\eta_n | \theta) d\eta_n, \label{eq:model3}\\
    \text{Model 4:} \;\; P(X_1, \ldots, X_N) &= \int_\Theta \left( \prod_{n=1}^N \int_H P(X_n | \eta_n) \, p(\eta_n | \theta) d\eta_n \,\right) p(\theta) \, d\theta, \label{eq:model4}\\
    \text{Model 5:} \;\; P(X_1, \ldots, X_N | \theta) &= \prod_{n=1}^N \int_H P(X_n | \eta_n, \theta) \, p(\eta_n | \theta) d\eta_n , \label{eq:model5}\\
    \text{Model 6:} \;\; P(X_1, \ldots, X_N) &= \int_\Theta \left( \prod_{n=1}^N \int_H P(X_n | \eta_n, \theta) \, p(\eta_n | \theta) d\eta_n \,\right) p(\theta) \, d\theta. \label{eq:model6}
\end{align}
For simplicity of notation, we are assuming that all random variables have densities. In the case of discrete random variables, the corresponding integrals are replaced by sums. In Figure 1, we use \textit{probabilistic graphical models} \citep{Koller} to describe the structure of these hierarchical models. In this figure, the plate (rectangle with subscript $N$) means that there are $N$ conditionally i.i.d. copies of the variables inside, given the variable outside. Gray nodes correspond to latent/unobserved variables, while white nodes correspond to observed/known variables. 

In models 1, 3, and 5, parameter $\theta$ is known/observed, thus we write  $P(X_1, \ldots, X_N | \theta)$. These three models describe i.i.d. (thus exchangeable) sequences, with increasingly complex hierarchical structures. Model 1 corresponds to a simple i.i.d. sequence, where each $X_n \sim P(X|\theta)$. In model 3, we have $X_1,...,X_n$ conditionally independent given $\eta_1,...,\eta_N$, with $X_n \sim P(X|\eta_n)$ and, in turn, the $\eta_n \sim p(\eta|\theta)$ conditionally i.i.d., given $\theta$. Consequently, $P(X_n|\theta)$ is given by the inner integral in Equation \eqref{eq:model3}. Finally, model 5 is an generalization of model 3 where
$X_n \sim P(X|\eta_n,\theta)$, i.e., $X_n$ is not conditionally independent of $\theta,$ given $\eta_n$. 

Models 2, 4, and 6 correspond to exchangeable (but not i.i.d.) sequences, as the outer variable $\theta$ in the hierarchy is not observed, thus it is marginalized out in Equations \eqref{eq:model2}, \eqref{eq:model4}, and \eqref{eq:model6}. Model 2 is simply the one addressed in the previous subsection, as Equation \eqref{eq:model2} coincides with Equation \eqref{definettiref}. Models 4 and 6 include an additional collection of random variables, with a structure similar to models 3 and 5, respectively.

\begin{figure}[hbt]
\centering
\begin{tikzpicture}[
    scale=0.65,
    transform shape,
    node distance=1.2cm,
    latent/.style={circle, draw=black, fill=gray!20, thick, minimum size=10mm},
    observed/.style={circle, draw=black, fill=white, thick, minimum size=10mm},
    arrow/.style={-Stealth, thick}
]

\begin{scope}[xshift=0cm]
    \node[observed] (theta2) {$\theta$};
    \node[latent, opacity=0] (eta2) [below=of theta2] {$\eta_n$};
    \node[observed] (x2) [below=of eta2] {$X_n$};
    \draw[arrow] (theta2) to [bend right=40] (x2);
    \begin{scope}[on background layer]
        \node[draw, thick, inner sep=6pt, fit=(x2)] (plate2) {};
        \node[anchor=south east] at (plate2.south east) {$N$};
    \end{scope}
    \node[below=1.2cm of x2, align=center] {Model 1\\ \textbf{i.i.d.}};
\end{scope}

\begin{scope}[xshift=2.8cm]
    \node[latent] (theta3) {$\theta$};
    \node[latent, opacity=0] (eta3) [below=of theta3] {$\eta_n$};
    \node[observed] (x3) [below=of eta3] {$X_n$};
    \draw[arrow] (theta3) to [bend right=40] (x3);
    \begin{scope}[on background layer]
        \node[draw, thick, inner sep=6pt, fit=(x3)] (plate3) {};
        \node[anchor=south east] at (plate3.south east) {$N$};
    \end{scope}
    \node[below=1.2cm of x3, align=center] {Model 2\\ \textbf{exchangeable}};
\end{scope}

\begin{scope}[xshift=5.6cm]
    \node[observed] (theta5) {$\theta$};
    \node[latent] (eta5) [below=of theta5] {$\eta_n$};
    \node[observed] (x5) [below=of eta5] {$X_n$};
    \draw[arrow] (theta5) -- (eta5);
    \draw[arrow] (eta5) -- (x5);
    \begin{scope}[on background layer]
        \node[draw, thick, inner sep=6pt, fit=(eta5) (x5)] (plate5) {};
        \node[anchor=south east] at (plate5.south east) {$N$};
    \end{scope}
    \node[below=1.2cm of x5, align=center] {Model 3\\ \textbf{i.i.d.}};
\end{scope}

\begin{scope}[xshift=8.4cm]
    \node[latent] (theta6) {$\theta$};
    \node[latent] (eta6) [below=of theta6] {$\eta_n$};
    \node[observed] (x6) [below=of eta6] {$X_n$};
    \draw[arrow] (theta6) -- (eta6);
    \draw[arrow] (eta6) -- (x6);
    \begin{scope}[on background layer]
        \node[draw, thick, inner sep=6pt, fit=(eta6) (x6)] (plate6) {};
        \node[anchor=south east] at (plate6.south east) {$N$};
    \end{scope}
    \node[below=1.2cm of x6, align=center] {Model 4\\ \textbf{exchangeable}};
\end{scope}

\begin{scope}[xshift=11.2cm]
    \node[observed] (theta7) {$\theta$};
    \node[latent] (eta7) [below=of theta7] {$\eta_n$};
    \node[observed] (x7) [below=of eta7] {$X_n$};
    \draw[arrow] (theta7) -- (eta7);
    \draw[arrow] (eta7) -- (x7);
    \draw[arrow] (theta7) to [bend right=45] (x7);
    \begin{scope}[on background layer]
        \node[draw, thick, inner sep=6pt, fit=(eta7) (x7)] (plate7) {};
        \node[anchor=south east] at (plate7.south east) {$N$};
    \end{scope}
    \node[below=1.2cm of x7, align=center] {Model 5\\ \textbf{i.i.d.}};
\end{scope}

\begin{scope}[xshift=14cm]
    \node[latent] (theta8) {$\theta$};
    \node[latent] (eta8) [below=of theta8] {$\eta_n$};
    \node[observed] (x8) [below=of eta8] {$X_n$};
    \draw[arrow] (theta8) -- (eta8);
    \draw[arrow] (eta8) -- (x8);
    \draw[arrow] (theta8) to [bend right=45] (x8);
    \begin{scope}[on background layer]
        \node[draw, thick, inner sep=6pt, fit=(eta8) (x8)] (plate8) {};
        \node[anchor=south east] at (plate8.south east) {$N$};
    \end{scope}
    \node[below=1.2cm of x8, align=center] {Model 6\\ \textbf{exchangeable}};
\end{scope}

\end{tikzpicture}
\caption{Graphical models representing the probabilistic models 1 to 6, defined in Equations \eqref{eq:model1} -- \eqref{eq:model6}. The plate represents $N$ i.i.d. copies of the random variable inside, conditioned on the one outside. The variables in white are observed, those in gray are not.}
\label{graphical_models1}
\end{figure}

The set of graphical models presented in Figure \ref{graphical_models1} illustrates that it is the ignorance versus knowledge about the outer variable $\theta$ (not its existence) that makes the sequence exchangeable but not independent (models 2, 4, and 6) versus i.i.d. (models 1, 3, 5). Furthermore, the presence (models 3, 4, 5, and 6) or absence (models 1 and 2) of the latent variables $\eta_1,...,\eta_N$ inside the plate has no effect of the independence or not of $X_1,...,X_n$.

We conclude this subsection with another urn example illustrating the hierarchical exchangeable model.

\vspace{0.1cm}

\begin{description}
    \item[Hierarchical exchangeable urn sequence example:] Consider two same two urns ($A$ and $B$) with different ratios of black and white balls to the total number of balls (denoted $\eta_{(A)}$ and $\eta_{(B)}$). Now, in each draw (rather than only once in the beginning of the process), a coin is flipped to select one of the urns (although the choice is not observed), and a ball is sampled from the selected urn, with replacement. This coin is one of two different coins, coin 1 and coin 2, selected before the sampling process begins by flipping a third, fair coin. Coin 1 selects urn A with probability $\theta_1$ (thus urn $B$ with probability $1-\theta_1$) and coin 2 selects urn A with probability $\theta_2$ (thus urn $B$ with probability $1-\theta_2$). Which coin is being used is not observed. This corresponds to a mixture of two i.i.d. processes, such that each sample of this process is itself also a mixture of two distributions (the two urns). 
    
   This process follows model 4 defined by Equation \eqref{eq:model4}, with the following definitions: $X_n\in \{0,\, 1\}$, corresponding to black and white balls, respectively, and $P(X_n|\eta_n) = \eta_n^{X_n} (1-\eta_n)^{(1-X_n)}$, for $\eta_n \in \{ \eta_{(A)}, \theta_{(B)}\}$. $P(\eta_{(A)} | \theta_1) = \theta_1$ and $P(\eta_{(A)} | \theta_2) = \theta_2$. Finally, $p(\theta_1) = p(\theta_2) = 1/2$.
\end{description}

\vspace{0.1cm}

Note that it is not the mere existence of latent variables that makes the problem exchangeable. If $\theta$ and $\psi$ can only take one value, the outer integrals can be dropped, and the sequence is i.i.d. by definition. In the example above, it would be equivalent to having a single coin with known probabilities. Simultaneously, it is not the existence of both urns, or the fact that the experimental procedure alternates between urns, that makes this sequence exchangeable. It is the fact that the probability of sampling from each urn is unknown (initial choice of coin) that makes it so.

\subsection{Previous Work Intersecting Causality and Exchangeability}
Currently, the concepts of causality and exchangeability overlap in two principal domains: the well-established field of causal inference, which fundamentally relies on exchangeability, and the recent work by Guo et al. that generalizes de Finetti’s theorem to the causal case and introduces a novel approach to causal discovery \citep{cdf}.

\subsubsection{Exchangeability in Average Treatment Effect Estimation}
In the experimental setting, the goal is often to  infer whether a variable has a causal effect on another,  knowing the reverse is impossible (e.g., one variable corresponds to an event that precedes the other) \label{inferenceref}. This is often assessed using randomized trials that can be viewed through the potential outcomes framework \citep{potentialoutcomes}. Suppose the goal is to assess the effect of a treatment ($T$) on the outcome of some disease ($Y$). The common approach is to gather a group of patients, of which a randomly selected half is subject to the treatment ($T=1$), and the other half (the control group) is not ($T=0$). The outcome ($Y\in\{0,\, 1\}$) is recorded for each patient. From this data, the goal is to estimate the so-called \textit{average treatment effect} (ATE), defined as 
\begin{equation}
    \text{ATE} = \mathbb{E}_D[Y(1) - Y(0)]=\mathbb{E}_D[Y(1)] - \mathbb{E}_D[Y(0)],
\end{equation}
where $\mathbb{E}_D$ denotes expected value over the entire distribution, and $Y(1)$ and $Y(0)$ are two random variables representing the outcome for $T=1$ and $T=0$ respectively \citep{Holland}. The main challenge lies in estimating this quantity (even in the presence of infinitely many patients) since, for each patient, either $Y(1)$ or $Y(0)$ is observed, never both. Additionally, it makes no sense to consider the patients as i.i.d. samples, given the relevant differences between them, such as age and health condition. 

To estimate the ATE, it is sufficient to assume exchangeability \citep{ate,rosenbaum1983central}, as reviewed next. Suppose the indices up to $N$ belong to the control group and those from $N+1$ to $2N$ to the treatment group, and that the goal is to generalize the results of the treatment group to the control group. Begin by noticing that exchangeability guarantees the following equality:
\begin{equation}
    P(Y_1(1), \ldots, Y_N(1), Y_{N+1}(1), \ldots, Y_{2N}(1)) 
    = P(Y_{N+1}(1), \ldots, Y_{2N}(1), Y_1(1), \ldots, Y_N(1)), 
\end{equation}
Marginalizing both sides of the equality with respect to the second half of the arguments yields
\begin{equation}
P(Y_1(1), \ldots, Y_N(1))  = P(Y_{N+1}(1), \ldots, Y_{2N}(1)),
\end{equation}
which implies that 
\begin{equation}
\mathbb{E}_{D}[Y(1)|T=0] = \mathbb{E}_{D}[Y(1)| T = 1] =  \mathbb{E}_D[Y(1)]. 
\end{equation}
Therefore, exchangeability implies that $\mathbb{E}[Y(1) | T=1]$ can replace $\mathbb{E}_D[Y(1)]$ and similarly for the control group ($\mathbb{E}[Y(0) | T=0]$ can replace $\mathbb{E}_D[Y(0)]$), which allows computing the ATE \citep{hernan2020causal}. More intuitively, exchangeability means that it is fair to generalize the results obtained in one group to the entire population. This is valid since the patients in both groups were drawn from the same latent distribution \citep{hofler2005causal}. 

Additionally, the converse is also true: if the data is not exchangeable, then it is impossible to estimate the ATE, since there is no valid argument to generalize to the entire population. For example, if there latent distributions of both groups are different (e.g., one group could be younger than the other). This would invalidate extrapolating the results obtained from one group to the other, which could lead to misleading results. Nevertheless, if one is aware of such imbalances, it is possible to adjust for them by considering them as observed variables (if they are observed), and performing the exact same reasoning while conditioning on these observed variables \citep{lee2022standardization}.

Interestingly, within the potential outcomes framework, exchangeability implies that correlation is equivalent to causation \citep{rosenbaum1983central,hernan2020causal}. This is because, under the exchangeability hypothesis, all latent variables (which could generate confounding) are controlled for by both groups following the same prior.

\subsubsection{The Causal de Finetti Theorem}
The connection between exchangeability and causal discovery was made only recently\label{cdfref}, when Guo et al. adapted de Finetti's Theorem (see Section \ref{definettiref}) to the setting of causal discovery \citep{cdf}. The theorem involves the notion of infinite exchangeability for an infinite sequence of pairs $(X_1,Y_1,...,X_N,Y_N, ...)$, which holds if, for any $N$ and any permutation $\pi:\{1,...,N\}\rightarrow\{1,...,N\}$,
\begin{equation}
    P(X_{\pi(1)},Y_{\pi(1)} \ldots, X_{\pi(N)},X_{\pi(N)}) = P(X_1,Y_1, \ldots, X_N,Y_N).
\end{equation}

\vspace{0.15cm}

\begin{description}
    \item[Causal de Finetti Theorem \citep{cdf}:] 
\textit{Let \(\{(X_n, Y_n)\}_{n \in \mathbb{N}}\) be an infinite sequence of binary random variable pairs. The sequence satisfies the two following properties
\begin{enumerate}
\item infinite exchangeability, and
\item \(\forall n \in \mathbb{N}: Y_{[n]} \perp\!\!\!\perp  X_{n+1} | X_{[n]}\), where $[n]=\{1,\ldots,n\}$, 
\end{enumerate} 
\vspace{-0.4cm} if and only if there exist two random variables \(\theta \in [0, 1]\) and \(\psi \in [0, 1]^2\), with probability measures \(\mu\) and \(\nu\), respectively, such that}
\begin{equation} \label{cdfequation}
    P(X_1, Y_1, \dots, X_N, Y_N) = \int_{\Phi} \int_{\Theta} \prod_{n=1}^N p(Y_n | X_n, \psi) p(X_n | \theta) \, d\mu(\theta) \, d\nu(\psi).
\end{equation} 
\end{description} 
In \eqref{cdfequation}, $d\mu(\theta)$ and $d\nu(\psi)$ can be substituted by $p(\theta)d\theta$ and $p(\psi)d\psi$, respectively, if $\mu$ and $\nu$ are absolutely continuous w.r.t. the Lebesgue measure on $\Theta$ and $\Psi$. Furthermore, as in the original de Finetti theorem, if both $\Theta$ and $\Phi$ are discrete sets, Equation \eqref{cdfequation} takes the form
\begin{equation} \label{cdfequation_disc}
    P(X_1, Y_1, \dots, X_N, Y_N) = \sum_{\phi\in\Phi} \sum_{\theta\in \Theta} \prod_{n=1}^N p(Y_n | X_n, \psi) p(X_n | \theta) \, p(\theta) \, p(\psi),
\end{equation} 
where $p(\theta)$ and $p(\psi)$ are probability mass functions. 

Guo et al. also argue that eschewing stronger i.i.d. constraints for exchangeability allows the presence of asymmetric dependencies between variables. These asymmetries can then be explored to develop new causal discovery methods. Specifically, they propose an algorithm for bivariate causal discovery based on the asymmetry between both causal directions present in the second assumption of the Causal de Finetti Theorem. \label{cdfalgorithm} While in the causal direction, $Y_{[n]} \perp\!\!\!\perp X_{n+1} | X_{[n]}$ holds, the same is not necessarily true in the opposite direction. To test whether \(\forall n \in \mathbb{N}: Y_{[n]} \perp\!\!\!\perp X_{n+1} | X_{[n]}\), one would need to have the joint probability distribution over $P(Y_{[n]}, X_{n+1}, X_{[n]})$. However, the asymmetry expressed above is impossible to test in an exchangeable process, since only one realization of each variable is obtained. To address this issue, the authors constructed datasets that approximate de Finetti's Theorem in viewing an exchangeable process as a mixture of i.i.d. sequences. Specifically, they assume the data comes from $N$ different environments, where each environment, $e$, is defined by its latent parameters, which are drawn according to the prior distribution: $(\theta^e,\psi^e) \sim(p(\theta),p(\psi))$. In a sense, this is similar to the earlier concept of interventions, because all environments share the same underlying causal graph over the observed variables $X$ and $Y$, while the specific causal mechanism is defined by $\theta$ and $\psi$. Nevertheless, from a more practical viewpoint, these interventions are much softer between each other (very similar causal relationships between interventions), and many more in number (the authors' algorithm considers 100).

\section{Exchangeability for Causal Discovery}

\subsection{Hierachical Model}\label{hier_section}
In this section, with the aim of offering a more complete and flexible description of causal processes, a new hierarchical model is proposed. This model extends the Causal de Finetti representation theorem (Equation \eqref{cdfequation}) in a similar way that the hierarchical models in Subsection \ref{sec:hier} extend the original de Finetti theorem (Equation \eqref{definettiref}). The proposed formulation takes the terms 
$p(X_n | \theta)$ and $p(Y_n | X_n, \psi)$ as representing two marginal probability distributions, integrated over inner latent variables.  With $\eta_n$ and $\gamma_n$ representing these latent variables, marginalization with respect to them yields
\begin{eqnarray}\label{hier_exc_eq}
    \lefteqn{P(X_1, Y_1, \dots, X_N, Y_N) } \nonumber\\ 
    & = & \int_{\Phi} \int_{\Theta} \prod_{n=1}^N \left( \int_{H}  p(X_n | \eta_n, \theta) \,p(\eta_n | \theta) \, d\eta_n \right) \left( \int_\Gamma p(Y_n | X_n, \gamma_n, \psi) \, p(\gamma_n | \psi) \, d \gamma_n \right) \,p(\theta) p(\psi) \,  d\theta \, d\psi,
\end{eqnarray}
where 
\begin{equation}\label{hier_pX}
     \int_{H}  p(X_n | \eta_n,\theta ) \,p(\eta_n | \theta) \, d\eta_n = p(X_n | \theta),
\end{equation}
and
\begin{equation}\label{hier_pY_X}
     \int_\Gamma p(Y_n | X_n, \gamma_n, \psi) \, p(\gamma_n | \psi) \, d \gamma_n = p(Y_n | X_n, \psi).
\end{equation}
As above, the integrals in these three equations are replaced by sums in the case of discrete variables. 

This hierarchical model provides more flexibility, while still allowing for efficient sampling.  The equations above are purposefully general: the variables ($\eta_n, \gamma_n, \theta, \psi$) can be scalars or vectors, continuous, discrete, or mixed, depending on the exact nature and formulation of the problem being addressed. Namely, the variable $\theta$ plays two roles in that it defines the distribution of $\eta_n$ and the conditional distribution of $X_n$ given $\eta_n$. The same holds for $\psi,$ which defines the distribution of $\gamma_n$ as well as the conditional distribution of $Y_n$, given $X_n$ and $\gamma_n$ (\textit{i.e.}, the causal mechanism). A graphical representation of this model, alongside the exchangeable model proposed by Guo et al. \citep{cdf}, is shown in Figure \ref{fig:three_dags}.

\begin{figure}[htbp]
\centering
\begin{tikzpicture}[
    scale=0.8,
    transform shape,
    node distance=1cm and 2.8cm,
    latent/.style={circle, draw=black, fill=gray!20, thick, minimum size=10mm},
    observed/.style={circle, draw=black, fill=white, thick, minimum size=10mm},
    arrow/.style={-Stealth, thick}
]

\begin{scope}

    \node[latent] (thetaM) {$\theta$};
    \node[latent] (psiM) [right=of thetaM] {$\psi$};

    \node[latent, opacity=0] (etaM) [below=of thetaM] {$\eta_n$};
    \node[latent, opacity=0] (gammaM) [right=of etaM] {$\gamma_n$};

    \node[observed] (xM) [below=of etaM] {$X_n$};
    \node[observed] (yM) [below=of gammaM] {$Y_n$};

    \draw[arrow] (thetaM) -- (xM);
    \draw[arrow] (psiM) -- (yM);
    \draw[arrow] (xM) -- (yM);

    \begin{scope}[on background layer]
        \node[draw, thick, inner sep=10pt,
              fit=(etaM) (xM) (gammaM) (yM)] (plateM) {};
        \node[anchor=south east] at (plateM.south east) {$N$};
    \end{scope}

\end{scope}

\begin{scope}[xshift=8.5cm]

    \node[latent] (thetaR) {$\theta$};
    \node[latent] (psiR) [right=of thetaR] {$\psi$};

    \node[latent] (etaR)   [below=of thetaR] {$\eta_n$};
    \node[latent] (gammaR) [right=of etaR] {$\gamma_n$};

    \node[observed] (xR)   [below=of etaR] {$X_n$};
    \node[observed] (yR)   [below=of gammaR] {$Y_n$};

    \draw[arrow] (thetaR) -- (etaR);
    \draw[arrow] (thetaR) to [bend right=45] (xR);
    \draw[arrow] (etaR)   -- (xR);

    \draw[arrow] (psiR)   -- (gammaR);
    \draw[arrow] (psiR)   to [bend left=45] (yR);
    \draw[arrow] (gammaR) -- (yR);

    \draw[arrow] (xR) -- (yR);

    \begin{scope}[on background layer]
        \node[draw, thick, inner sep=10pt,
              fit=(etaR) (xR) (gammaR) (yR)] (plateR) {};
        \node[anchor=south east] at (plateR.south east) {$N$};
    \end{scope}

\end{scope}

\end{tikzpicture}

\caption{Left: exchangeable model of Guo et al. \citep{cdf}. Right: proposed exchangeable hierarchical model. The plate represents N i.i.d. samples, while the variables in white are observed, while those in grey are not.}
\label{fig:three_dags}
\end{figure}

It is important to note that $p(Y_n|X_n,\psi)$ in Equation \eqref{hier_pY_X} can entail two fundamentally different types of interactions. Analogously to the previous example, in the term $p(\gamma_n|\psi)$, $\psi$ defines the shape and parameters of the inner latent variable $\gamma_n$ (e.g., if $\psi<0$, $p(\gamma_n|\psi) = \mathcal{N}(0,1) \, \text{; if} \, \psi>0, p(\gamma_n|\psi) = \mathcal{U}(0,1)$). On the other hand, $\psi$ can also select (or inform) over the causal mechanism between $X_n$ and $Y_n$ (e.g., if $\psi$ is smaller than 0, then the $Y_n = X_n^2$, otherwise  $Y_n \sim \mathcal{B}(X_n,\psi^2)$).
These two different and complementary interpretations of Equation \eqref{hier_exc_eq} allow for the proposal of two factors that should lead causal sequences to be considered exchangeable and not i.i.d.: 1) when the causal mechanism between $X_n$ and $Y_n$ is unknown, and 2) when there exist latent variables whose distribution is unknown. This stems from the fact that the lack of knowledge over either of these two components can be accurately modeled using an exchangeable sequence and, more precisely, the hierarchical model presented in this section.

\subsection{Arguments}

The key contribution of this work is the proposal that i.i.d. causal discovery is more accurately framed as exchangeable, in accordance with the hierarchical model defined in Subsection \ref{hier_section}. According to this model, it suffices for either the causal mechanism between $X_n$ and $Y_n$, or the distribution of any existing latent variables, to be unknown to break the i.i.d. assumption. In fact, we believe it is reasonable to argue that this is most often the case in real data. Bivariate real-world examples tend to stem from very complex causal networks, where both mechanisms and latent variables are not easy to identify, often being left as underlying assumptions. In fact, if only considering the existence of latent variables with an unknown distribution, an analysis of the Tübingen collection led to the conclusion that $74.4\%$ of datasets are more accurately described as exchangeable, while $10.3\%$ have no clear latent variables with unknown distribution, and $15.3\%$ are time series (see Appendix \ref{Tübingen_appendix} for a full breakdown). These statistics in the Tübingen collection, the only real-world bivariate causal discovery dataset, further consolidate the core claim of this work.

An additional argument stems from the clear similarities between causal inference and causal discovery. Crucially, they share the same setup and the same ultimate goal. The difference is that, in causal inference, it is possible to intervene directly in the system, while causal discovery relies solely on observational data. Since they operate under the same formal assumptions and objectives, it makes sense to require the same foundational properties of the data in both settings. In experimental causal inference, one of the key assumptions is exchangeability (see Section \ref{inferenceref}). It enables controlling for latent factors without necessitating the assumption of their absence, which is essential for addressing the complexities of real-world data. As such, the exchangeability assumption in causal inference has been widely accepted and thoroughly validated in the scientific community as the criterion that enables extrapolating causal conclusions drawn from experimental data \citep{ate}. While experimental causal inference has long been a routine, rigorously validated practice, observational causal discovery remains primarily an exploratory research field. Consequently, observational causal discovery should look towards causal inference in order to extrapolate verified and meaningful assumptions from the data. More specifically, it should resemble causal inference in considering exchangeability to be a key statistical property of observational data. 

The last argument will rely on the expressiveness of a synthetic dataset generated in accordance with the hierarchical model defined in Equation \eqref{hier_exc_eq}. By relaxing the stricter i.i.d. assumption, the hierarchical model enables the synthetic generation process to encompass a broader family of causal mechanisms, better reflecting the inherent complexities of real-world data. Having the Tübingen dataset as a baseline, we will first discuss how different methods score more closely (in comparison with the Tübingen) relative to other synthetic datasets (Section \ref{synthetic dataset section}). Then, in Section 5, we will introduce a new causal discovery method: a neural network trained exclusively on the developed synthetic dataset, and show how it performs similarly to other state-of-the-art methods (Section \ref{synthNN_start}).

\section{Synthetic Dataset}\label{synthetic dataset section}

\subsection{Motivation and Objectives}
This section describes the creation of a synthetic dataset of cause-effect pairs, generated by following the hierarchical version of the Causal de Finetti representation theorem, expressed in Equation \eqref{hier_exc_eq}. The creation of this synthetic dataset aims to fulfill the two following objectives simultaneously:
\begin{itemize} \label{synth_objectives}
    \item Showcasing the central role of exchangeability in causal discovery, achieved by generating samples according to the hierarchical version of the Causal de Finetti theorem in Equation \eqref{hier_exc_eq}. By doing so, if this dataset is able to match real-world dependency structures, it serves as further confirmation that exchangeable (and not necessarily i.i.d.) distributions are a core concept for causal discovery.
    \item Providing an additional, valid dataset for testing new causal discovery methods. As noted in Subsection \ref{Tübingen_intro}, the Tübingen dataset remains the sole real-world benchmark for bivariate causal discovery. Existing synthetic datasets, in contrast, are used only to evaluate method validity under particular assumptions. The goal of our dataset is to offer a properly constructed synthetic resource for method analysis. We propose using it complementarily alongside the Tübingen dataset, with its synthetic nature allowing for far greater precision (since arbitrary numbers of examples can be generated as needed).
\end{itemize}
To guarantee both objectives above, it is necessary to demonstrate that the dataset is valid. In other words, it must be shown to be representative of real-world scenarios. Because the only available real-world reference is the Tübingen dataset, this can be achieved by demonstrating that various state-of-the-art methods produce very similar results on both the Tübingen dataset and the proposed synthetic dataset.

\subsection{Algorithm}

The algorithm for synthetic dataset generation is based on the proposed hierarchical exchangeable model in Equation \eqref{hier_exc_eq}. Firstly, to ensure consistency of orders of magnitude between variables, $X_n,Y_n,\eta_n,$ and $\gamma_n$ were always subjected to min-max scaling after sampling. This aims to minimize a method's ability to artificially explore asymmetries between the two causal directions, arising from designer biases in the ranges of values that the different variables take.

In accordance with Equation \eqref{hier_exc_eq}, and the subsequent interpretation, $\theta$ and $\psi$ will represent both the shape and the parameters of the unknown distributions of the inner latent variables $\eta_n,\gamma_n$, and the choice of functional mechanisms (how each $X_n$ depends on $\eta_n$, and how each $Y_n$ depends on $X_n$ and $\gamma_n$, respectively). In this generation algorithm, these two possible contributions of $\theta$ and $\psi$ are decoupled. Therefore, Equations \eqref{hier_pX} and  \eqref{hier_pY_X} are rewritten by considering $\theta$ and $\psi$ to be two vectors $\theta=[\theta_1,\theta_2];\, \psi=[\psi_1,\psi_2]$, where each subvector controls one of the two possible impacts of $\theta$ and $\psi$, respectively:
\begin{align}\label{variable_split_eq}
    \int_{H}  p(X_n | \eta_n,\theta_2) \,p(\eta_n | \theta_1) \, d\eta_n = p(X_n | \theta), \, \\
     \int_\Gamma p(Y_n | X_n, \gamma_n, \psi_2) \, p(\gamma_n | \psi_1) \, d \gamma_n = p(Y_n | X_n, \psi).
\end{align}

Furthermore, the hierarchical model in Equation \eqref{hier_exc_eq} does not place any constraints on the values $\theta$ and $\psi$ can take. Nevertheless, in our algorithm, since $\theta_1$ and $\psi_1$ will represent the choice of the inner latent variables distribution, they were sampled from a uniform discrete distribution over the available distributions. This choice is an algorithm's hyperparameter, and we opted for the following three possible distributions: Uniform, Gaussian, and Rayleigh. These distributions were chosen given their ubiquitous presence in different applications of statistics. Additionally, these distributions have the significant advantage that their parameters are invariant to min-max scaling. Therefore, there is no need to select the distribution's parameters. 

Variables $\theta_2$ and $\psi_2$  control the conditional probabilities of $X_n$ and $Y_n$. In the case of $\theta_2$, all information regarding $X_n$ can be directly controlled for by the shape of the distribution of $\eta_n$ (enforced by $\theta_1$). Therefore, this term is redundant and will not be taken into consideration in the presented algorithm. In the case of $\psi_2$, it controls the causal mechanism from $X_n, \gamma_n$ to $Y_n$. It simultaneously represents the sampled mathematical family of functional mechanisms and the specific function's parameters (the combination of these two will be denoted $f_{\psi_2}$).

Regarding the mathematical family of functions representing the causal relationship, it is uniformly sampled from the following 1) linear; 2) piecewise linear; 3) exponential; 4) logarithm; 5) inversely proportional; 6) Brownian-like motion\footnote{Note that this function, despite requiring knowledge of previous samples to compute the latter, does not constitute a time series, given there is no underlying time variable. In fact, it only aims at generating a differentiable function in its most general form.}; 7) polynomial; 8) power law. Additionally, regarding the function's parameters, these are sampled in accordance with fixed hyperparameters and sampling distributions that depend on the specific function. For further details, see Appendix \ref{algorithm_extras}. 

Lastly, two additional terms are included to represent the measurement noise of $X_n$ and $Y_n$. These are two Gaussian distributions with very small variances ($\sigma^2_X$ and $\sigma^2_Y$), to ensure they have little impact on the generation process. Thus, the conditional probabilities $p(X_n | \eta_n,\theta_2)$ and $p(Y_n | X_n, \gamma_n, \psi_2)$ take the form
\begin{align}
    p(X_n | \eta_n,\theta_2) = \mathcal{N}(\eta_n,\sigma^2_X),\\
    p(Y_n| X_n, \gamma_n, \psi_2 ) = \mathcal{N}\left(f_{\psi_{2}} (X_n, \gamma_n), \sigma^2_Y\right).
\end{align}
    
To generate a collection of pairs $(X_1,Y_1),...,(X_N,Y_N)$, $\theta$ and $\psi$ are sampled once, at the beginning of the generation process, in accordance with the models in Figure \ref{fig:three_dags}. Then, the $N$ samples of $\eta_n, \gamma_n, X_n$, and $Y_n$ are obtained as described in Algorithm \ref{gen_alg}. The value $N$ is sampled from a distribution that aims to mimic the sample size distribution in the Tübingen dataset. This is obtained by fitting a Gaussian mixture with 3 components to the data.
 
\begin{algorithm}
\caption{Synthetic Dataset Generation Algorithm}\label{gen_alg}
\hspace*{\algorithmicindent} \textbf{Input:} $\{P_{\theta_1}\}, \{P_{\psi_1}\}, \{f_{\psi_2}\}, N, \sigma^2_X,\sigma^2_Y$\\
\hspace*{\algorithmicindent} \textbf{Output:} $N$ causally related exchangeable pairs $\{(x_i,y_i)\}$
\begin{algorithmic}[1]
\State Choose uniformly from $\{P_{\theta_1}\}, \{P_{\psi_1}\}, \{f_{\psi_2}\}$.
\State $\eta_1,...,\eta_N \sim P_{\theta_1}, \mbox{i.i.d.}$
\State $\gamma_1,...,\gamma_N \sim P_{\psi_1}, \mbox{i.i.d.}$
\State Perform min-max scaling on both $(\eta_1,...,\eta_N)$ and $(\gamma_1,...,\gamma_N)$
\State For $n=1$ to $N,\; X_n \sim \mathcal{N}(\eta_n,\sigma^2_X)$ 
\State For $n=1$ to $N,\; Y_n \sim \mathcal{N}(f_{\psi_2}(X_n,\gamma_n),\sigma^2_Y)$
\State Perform min-max scaling on both $(X_1,...,X_N)$ and $(Y_1,...,Y_N)$.
\end{algorithmic}
\end{algorithm}

All considered mathematical descriptions of the causal mechanism (controlled by $\psi_2$) are designed to be strictly increasing in the latent variable ($\gamma_n$). Although this is not directly enforced by the theory of exchangeability, it constitutes a design choice based on theoretical and practical reasons. On the one hand, monotonic SCMs have been proposed as a reasonable simplification of the problem of causal discovery \citep{monotonicscms}; on the other hand, many real-world examples in the Tübingen collection (see Appendix \ref{Tübingen_appendix}) reinforce this belief, such as the relationship between age and height, or monthly rent and size in square meters\footnote{In the first example, the latent variable can be considered to be the genes; in this case, someone with "taller" genes may be both taller than other people at age 10, 40, 70, and so on. In the second example, considering a latent variable such as location, it is also reasonable to assume that independently of the size in square meters of an apartment, the better its location, the more expensive the price.}.

\subsection{Constructing the Full Dataset}
\label{synth_construct}
Each run of the algorithm described in the previous subsection generates a syntetic cause-effect example. Each example is composed of N samples, and obtained according to a single realization of the outer variables ($\theta$ and $\psi$). In this subsection, we will design a weighting scheme where the examples are weighted differently, based on similarity criteria with certain desired statistical properties. In our case, we will be aiming to mimic the statistics of the T\"ubingen dataset, and thus of real-world causal systems. To do so, they must depend on the realized value of the outer variables in each example. Specifically, we will be looking at the shape of the distribution of the inner latent variables (defined by $\theta_1$ and $\psi_1$) and the chosen family of functions to represent the causal mechanisms (defined by $\psi_2$). For simplicity, we will not be taking into consideration the specific function parameters.

Alternatively, this process can also be seen as an attempt to model the prior distributions of $\theta_1$, $\psi_1$, and $\psi_2$ to better match observed patterns. Nevertheless, weighting the different combinations of outer variables allows for more flexibility, since different sets of weights can be used based on both differences in applications and differences in similarity criteria.



To help in estimating the distributions of the exchangeable variables, nine different established causal discovery methods were implemented\footnote{The implementations of RECI and IGCI were taken verbatim from Kalainathan and Goudet \citep{causalcode}. CGNN and ANM are based on the same repository, with minor adjustments to their hyperparameters (for ANM, we replaced its original independence test with the Hoeffding’s D test to improve computational efficiency). The bQCD and EMD methods were sourced from Tagasovska et al., and SLOPE was implemented as described in its original publication \citep{bcqd,slope}. Finally, the LiNGAM and PNL algorithms were obtained from the GitHub repository at \url{https://github.com/ssamot/causality/tree/master}.}: ANM \citep{peters2010identifying}, bQCD \citep{bcqd}, CGNN \citep{cgnn}, EMD \citep{rkhs}, IGCI \citep{mian2023information}, LiNGAM \citep{shimizu2006linear}, PNL \citep{zhang2010distinguishing}, RECI \citep{reci}, and SLOPE \citep{slope}. Their performance in two different metrics, AUROC (area under the ROC curve) and accuracy (the two metrics are most widely present in the literature), was recorded for each combination of the three relevant exchangeable variables. In Table \ref{across_functions}, marginalized results over uniform choices of $\theta_1,\psi_1$ can be seen for these nine methods.

\begin{table}[h!]
\centering
\resizebox{\linewidth}{!}{
\begin{tabular}{lcccccccc}
\toprule
\textbf{Method }      & \textbf{All}   & \textbf{Brownian-like} & \textbf{Exponential}   & \textbf{Inv. Proportional}   & \textbf{Linear} & \textbf{Logarithmic}   & \textbf{Piecewise}   & \textbf{Polynomial} \\
\midrule
\textbf{ANM}         & 0.529 (0.512) & 0.248 (0.296) & 0.768 (0.703) & 0.256 (0.251) & 0.538 (0.514)  & 0.785 (0.771) & 0.232 (0.277) & 0.194 (0.256) \\
\textbf{bQCD}        & 0.420 (0.421) & 0.500 (0.494) & 0.463 (0.453) & 0.570 (0.531) & 0.354 (0.365)  & 0.246 (0.286) & 0.468 (0.456) & 0.573 (0.511) \\
\textbf{CGNN}        & 0.682 (0.653) & 0.635 (0.624) & 0.777 (0.722) & 0.727 (0.673) & 0.583 (0.565)  & 0.854 (0.810) & 0.605 (0.591) & 0.559 (0.562) \\
\textbf{EMD}         & 0.737 (0.718) & 0.667 (0.655) & 0.964 (0.906) & 0.811 (0.785) & 0.551 (0.574)  & 0.969 (0.938) & 0.559 (0.575) & 0.584 (0.575) \\
\textbf{IGCI}        & 0.758 (0.726) & 0.704 (0.675) & 0.952 (0.896) & 0.836 (0.789) & 0.594 (0.599)  & 0.986 (0.939) & 0.583 (0.593) & 0.588 (0.592) \\
\textbf{LiNGAM}      & 0.508 (0.515) & 0.466 (0.462) & 0.644 (0.612) & 0.567 (0.574) & 0.315 (0.365)  & 0.756 (0.701) & 0.363 (0.419) & 0.403 (0.432) \\
\textbf{PNL}         & 0.478 (0.491) & 0.533 (0.531) & 0.452 (0.485) & 0.487 (0.496) & 0.473 (0.487)  & 0.449 (0.477) & 0.495 (0.492) & 0.487 (0.480) \\
\textbf{RECI}        & 0.742 (0.732) & 0.678 (0.686) & 0.867 (0.876) & 0.851 (0.821) & 0.611 (0.630)  & 0.975 (0.941) & 0.615 (0.619) & 0.614 (0.596) \\
\textbf{SLOPE}       & 0.757 (0.729) & 0.683 (0.665) & 0.967 (0.922) & 0.828 (0.803) & 0.610 (0.631)  & 0.953 (0.906) & 0.624 (0.642) & 0.609 (0.600)\\
\bottomrule
\end{tabular}
}
\caption{Marginalized AUROC (accuracy) over uniform choices of $\theta_1, \psi_1$ achieved by different state-of-the-art methods.}
\label{across_functions}
\end{table}

Given the objective of representing real-world causal systems, the weights of the distributions of the exchangeable variables ($\theta_1,\psi_1$ and $\psi_2$) were chosen so that the difference in performance for each method between the generated synthetic dataset and that of the T\"ubingen was minimized. Moreover, two different weight combinations are provided in our implementation: one aimed at minimizing the difference in performance with respect to AUROC, the other with respect to accuracy. Both were computed by framing the problem using least squares optimization. In other words, it consists of finding the weight combination that minimizes the sum of the squared difference between each method's performance in both datasets. Additionally, an $\ell_2$ regularizer was used to improve performance for unseen methods \footnote{The $\ell_2$ regularizer and its weight ($1$) were chosen to optimize the results of SynthNN (see Section \ref{synthNN_start}). In other words, the performance of SynthNN was used as a proxy for the representativeness of the synthetic dataset.}. This can be translated into the following mathematical expression:
\begin{equation} \label{weights_eq}
\begin{aligned}
\min_{w}\; & \|A\,w - b\|_2^2 \;+\; \,\|w\|_2, \\
\text{subject to}\; & w \ge 0, \\
                    & \mathbf{1}^\top w = 1,
\end{aligned}
\end{equation}

\noindent where $A\in\mathbb{R}^{m\times d}$ is a matrix where element $A_{i,j}$ corresponds to the performance of method $i$ in the generated dataset according to the combination of exchangeable variables $j$; $b\in\mathbb{R}^m$ is a vector containing the performance of the assessed methods in the Tübingen dataset; and $w\in\mathbb{R}^d$ is a vector representing the minimization variable: the corresponding weights of each exchangeable variables combination.

Lastly, AUROC is a ranked choice metric, which means it analyses the order of confidence between different guesses. Therefore, the equation above does not exactly optimize the average error between datasets in the AUROC metric. The weighted average of the AUROC for each combination of exchangeable variables is not the AUROC of the weighted dataset. Nevertheless, optimizing the AUROC exactly is a combinatorial problem, much harder to solve, making the expression above a useful and simple approximation.

\subsection{Analysis}

In order for the two objectives laid out initially to be fulfilled, the causal discovery algorithms should perform similarly in the synthetic dataset presented in this paper and in the Tübingen dataset.

\begin{figure}[h] 
    \centering
    \includegraphics[width=\linewidth]{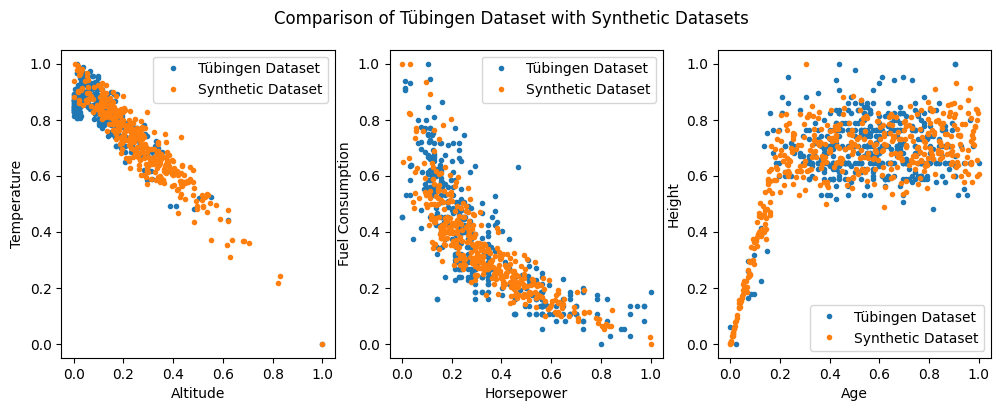}
    \caption{Three normalized Tübingen pairs plotted alongside hyperparameter-tuned examples of the synthetic dataset proposed in this paper.}
    \label{synth_vs_tueb}
\end{figure}

Firstly, it is important to understand which shapes the different generated examples can take. The different causal functions were designed to be as general and representative of the real world as possible. Figure \ref{synth_vs_tueb} shows how synthetic examples can match those in the Tübingen dataset by fixing the choice of function parameters. It seems reasonable to conclude that the data‑generation mechanism is sufficiently expressive to capture the variety of relationships found in the Tübingen dataset. Nevertheless, despite the ability of the algorithm to generate examples mimicking those in the Tübingen dataset, the inherent randomness present in the process ensures the actual examples are, in practice, more different (a random selection of these examples can be seen in Appendix \ref{examples_appendix}). It is also important to note that despite the apparent ability to represent a variety of causal functions present in the real-world, the Tübingen dataset has examples with discrete, categorical, or multi-dimensional variables, which are not present in our synthetic dataset.

Even though the proposed data generation process has the ability (through fine-tuning) to capture real-world examples, this is not enough to guarantee the quality of the dataset. For one, even though the generation process clearly passes the human visual test in terms of expressibility, this does not ensure it provides a proper representation of the causal element inherent to the real-world data. Secondly, even if the algorithm was demonstrably general enough to represent the causal structure of known real-world pairs via fine-tuning, this would not necessarily extrapolate to randomly generated examples. In other words, the clear causal element present in the fine-tuned examples might be a consequence of the fine-tuning itself, rather than the merits of the data generation algorithm.

A more comprehensive evaluation of the generated dataset was therefore undertaken. Nine established causal‐discovery algorithms were tested on both the Tübingen and our dataset, with performance measured by AUROC and accuracy. To facilitate a broader comparison, these methods were also applied to five additional synthetic datasets (including a noisy variant of our own). The full set of results is presented in Table \ref{methods_stats}.

\begin{table}[h]
\centering
\small
\begin{tabular}{ll|cccccccc}
\toprule
\multirow{2}{*}{Method} & \multirow{2}{*}{Metric} & \multicolumn{2}{c}{Ours} & Tübingen & \multicolumn{4}{c}{CE-datasets} \\
\cmidrule(lr){3-4} \cmidrule(lr){6-9}
 &  & Original & Noisy &  & Gauss & Net & Cha & Multi \\
\midrule
\multirow{2}{*}{\textbf{ANM}} & AUROC    & 46.4 (44.0) & 45.6 (45.1) & 44.7 & 12.1 & 21.0 & 35.2 & 74.0 \\
                          & accuracy & 41.0       & 40.2       & 39.7 & 19.3 & 29.3 & 37.7 & 61.3 \\
\addlinespace
\multirow{2}{*}{\textbf{CGNN}} & AUROC    & 68.9 (67.1) & 66.6 (64.9) & 66.7 & 70.6 & 67.1 & 62.5 & 93.6 \\
                              & accuracy & 61.8       & 61.0       & 62.5 & 64.7 & 62.0 & 61.3 & 85.0 \\
\addlinespace
\multirow{2}{*}{\textbf{EMD}} & AUROC    & 78.4 (70.3) & 80.4 (72.1) & 69.3 & 57.3 & 71.3 & 59.3 & 98.1 \\
                              & accuracy & 63.4       & 66.8       & 61.7 & 54.7 & 65.0 & 56.3 & 91.0 \\
\addlinespace
\multirow{2}{*}{\textbf{IGCI}} & AUROC    & 80.8 (75.1) & 80.4 (73.2) & 70.7 & 43.0 & 58.9 & 56.9 & 97.8 \\
                              & accuracy & 68.5       & 67.6       & 65.1 & 44.7 & 55.3 & 55.0 & 92.3 \\
\addlinespace
\multirow{2}{*}{\textbf{LiNGAM}} & AUROC    & 49.6 (50.7) & 50.8 (50.7) & 50.0 & 29.9 & 63.4 & 45.0 & 33.9 \\
                                & accuracy & 50.5       & 51.6       & 49.2 & 36.3 & 61.3 & 46.0 & 40.3 \\
\addlinespace
\multirow{2}{*}{\textbf{PNL}} & AUROC    & 47.1 (47.9) & 46.3 (46.0) & 41.3 & 44.6 & 51.4 & 48.1 & 42.2 \\
                              & accuracy & 48.2       & 49.3       & 44.5 & 45.7 & 49.7 & 48.0 & 45.7 \\
\addlinespace
\multirow{2}{*}{\textbf{bQCD}} & AUROC    & 61.3 (63.5) & 58.7 (61.1) & 73.0 & 50.9 & 92.2 & 58.7 & 56.6 \\
                               & accuracy & 60.2       & 57.1       & 70.0 & 55.3 & 84.0 & 58.3 & 50.3 \\
\addlinespace
\multirow{2}{*}{\textbf{RECI}} & AUROC    & 78.6 (74.9) & 78.2 (75.4) & 73.9 & 76.5 & 62.8 & 57.7 & 95.4 \\
                               & accuracy & 70.6       & 70.5       & 70.2 & 67.7 & 55.7 & 55.0 & 88.0 \\
\addlinespace
\multirow{2}{*}{\textbf{SLOPE}} & AUROC    & 82.1 (76.4) & 82.3 (76.0) & 78.9 & 73.1 & 66.9 & 59.4 & 96.9 \\
                                & accuracy & 70.6       & 69.8       & 71.5 & 67.3 & 62.3 & 57.0 & 88.7 \\
\bottomrule
\end{tabular}
\caption{Performance of different causal discovery methods (in terms of AUROC and accuracy) in the proposed syjthetic dataset, the Tübingen dataset, and five additional synthetic datasets. In the "Ours" datasets, the extra AUROC in parentheses refers to the weighted average AUROC obtained from Equation \eqref{weights_eq}}
\label{methods_stats}
\end{table}

In order to compare the results shown in Table \ref{methods_stats}, the average $\ell_1$ and $\ell_2$ distances to the Tübingen dataset were computed (which can be seen in Table \ref{synth_distances}). However, since this metric was used for computing the optimal weight distribution for this paper's dataset, it would be unfair to just present this metric, given the clear circularity in the argument: 1) The dataset quality can be assessed by how similar different methods perform when compared with the real-world reference (the Tübingen dataset) 2) The distribution of the exchangeable variables in the generation process was tuned to ensure that a selection of nine different causal discovery methods have very similar performances in both datasets. 3) Since these nine causal discovery methods have very similar performances in both datasets, the data-generation algorithm (and the consequent synthetic dataset) is shown to be very good. By finetuning the dataset using the performances of causal discovery methods, there is clear confounding in then using the same methods to assess the general similarity in performance for all causal discovery methods in the two datasets. Therefore, to control for this circularity, the average leave-one-out cross-validated $\ell_1$ and $\ell_2$ distances for this paper's datasets were also computed. It boils down to (one by one) first computing the weights by considering all but one method, and then analyzing the difference in performance only on the left-out method.

\begin{table}[h]
  \centering
  \resizebox{\textwidth}{!}{%
    \begin{tabular}{cc|cc|cc|cccc}
      \hline
      \multirow{2}{*}{\textbf{Norm}} & \multirow{2}{*}{\textbf{Metric}} 
        & \multicolumn{2}{c|}{\textbf{Ours (Original)}} 
        & \multicolumn{2}{c|}{\textbf{Ours (Noisy)}} 
        & \multicolumn{4}{c}{\textbf{CE-Datasets}} \\
      & 
        & Produced & CV & Produced & CV
        & Gauss & Net & Cha & Multi \\
      \hline
      \multirow{2}{*}{\centering $\ell_1$} 
        & AUROC     
        & 0.0544 (0.0297) & \textbf{0.0825} (0.0613) 
        & 0.0551 (0.0322) & 0.0809 (0.0627)
        & 0.1444          & 0.1150          & 0.1104          & 0.2056 \\
        & accuracy  
        & 0.0256         & \textbf{0.0519}          
        & 0.0351         & 0.0623
        & 0.0949         & 0.0878          & 0.0743          & 0.1836 \\
      \hline
      \multirow{2}{*}{\centering $\ell_2$} 
        & AUROC     
        & 0.0044 (0.0018) & \textbf{0.0169} (0.0115)
        & 0.0053 (0.0021) & 0.0156 (0.0107)
        & 0.0326          & 0.0180          & 0.0146          & 0.0496 \\
        & accuracy  
        & 0.0014         & \textbf{0.0083}
        & 0.0026         & 0.0104
        & 0.0144         & 0.0098          & 0.0082          & 0.0405 \\
      \hline
    \end{tabular}%
  }
  \caption{Performance of different synthetic datasets measured by comparing the average $\ell_1$ or $\ell_2$ norms of the difference between the achieved results (AUROC or accuracy) in the respective synthetic dataset and the Tübingen dataset.}
  \label{synth_distances}
\end{table}

Finally, as can be seen in Table \ref{synth_distances}, the presented dataset performs consistently better than four current synthetic datasets present in the literature (CE-Gauss, CE-Net, CE-Cha, and CE-Multi). The cross-validated original version of the synthetic dataset surpasses all other datasets in the $\ell_1$ metrics (AUROC: $0.0775$ and accuracy: $0.0516$) and is only nearly outperformed by the CE-Cha dataset in the $\ell_2$ metrics (AUROC: $0.0120$ and accuracy: $0.0062$). However, CE-Cha was built for the cause-effect pairs challenge and contains both artificial and real-world data from the Tübingen dataset, which causes clear confounding in this analysis. Furthermore, in absolute terms, the average difference in performance between our dataset and the Tübingen is also quite small. In all, it is clear that the presented synthetic dataset correctly mimics the Tübingen and consequently, real-world dynamics accurately. It is also clear that it does so to a higher degree than all other known synthetic datasets.

Additionally, the "noisy" dataset was constructed by applying significant additive and multiplicative noise to the original version, and it yields results that are marginally inferior to those of its unaltered counterpart. This suggests that there is nothing inherently causal in the presence of noise that cannot be captured appropriately by the underlying exchangeability in the dataset. This is even more interesting given the absence of any additive or multiplicative relationship between the cause and the inner latent variable ($\gamma_n$) for all implemented causal functions (see Appendix \ref{algorithm_extras}).

These findings become all the more striking when we recall that exchangeability was the only design principle guiding the construction of this dataset. Even more, despite being able to achieve very similar examples to the continuous ones in the Tübingen dataset, its actual generated examples are quite different (as can be seen in Appendix \ref{examples_appendix}). This is especially the case for variables that are not continuous or one-dimensional. Consequently, the presented dataset successfully fulfills its two initial objectives (see Subsection \ref{synth_objectives}):
\begin{itemize}
    \item It is clear this dataset resembles more closely the real-world than all other known synthetic datasets (see Table \ref{synth_distances}). Therefore, we hope its development will aid in better classifying, assessing, and comparing different causal discovery methods. More specifically, its complementary use may allow to compensate for some shortcomings of the Tübingen dataset, especially allowing for a more precise and general evaluation. 
    \item By allowing exchangeability to drive its development, we have produced a very strong synthetic dataset to date. 
    This success lends weight to the central thesis of our paper: that causal discovery under the i.i.d. assumption should, in fact, be reframed in terms of exchangeability.
\end{itemize}

\section{SynthNN}\label{synthNN_start}

Having created the synthetic dataset, as explained in Section \ref{synth_objectives}, a new causal discovery method named SynthNN is proposed, consisting of a neural network trained on the generated data. In each example, tabular data is provided, containing unordered causal pairs generated by Algorithm \ref{gen_alg}. The unordered nature of the pairs stems from the definition of exchangeability. Consequently, it is not clear at first how the input data should be fed into the neural network. Therefore, four different strategies were tested: 
\begin{enumerate}
    \item Feeding the data as a flattened one-dimensional vector into a fully connected layer: This is equivalent to ignoring all structure existing in the data. Essentially, the relationship between each $(x_i,y_i)$ pairing is not enforced. This means the network has to learn (in whichever way it finds appropriate) the structure during training. 
    \item Perform feature extraction for each pair, pool all the features together, and process this feature vector; this approach is based on the PointNet architecture developed for 3D point cloud classification \citep{pointnet}. Essentially, a shared MLP is applied individually to each pair $(x_i,y_i)$, which generates a higher order feature vector. Then, a global \mbox{max}pooling is applied to this vector, which can be interpreted as keeping the most relevant aspects of each pair and, consequently, of the input data overall. Finally, this global feature vector is processed by another MLP.
    \item Start by constructing a graph and then applying methods from graph neural networks (GNN) to the problem at hand. Specifically, the graph is constructed using k-nearest neighbours; then, several graph convolutional layers are applied according to the implementation by Kipf and Welling \citep{gcn}. Afterwards, the extracted features are flattened and processed by an MLP.
    \item Building an image from the data and processing it using a convolutional neural network (CNN). This image corresponds exactly to the images shown when plotting the generated data. Out of the four, this is the only method that had been previously applied to the problem of classifying cause-effect pairs by Singh et al. \citep{cnntueb}. Therefore, the designed network is generally based on this earlier implementation.
\end{enumerate}
At first look, the second and fourth methods have a clear advantage: they are inherently invariant in the number of pairs, which is crucial to the classification problem at hand. Alternatively, the other two would require padding so that the dimensions of all examples match. However, after implementing the four different methods, the fourth, based on image processing, clearly stood out as the most promising one. Consequently, the others were dropped, and the focus will be on this one. The full details of SynthNN, the implemented neural network, are in Appendix \ref{nn_specs}.

\begin{figure}[h] 
    \centering
    \includegraphics[width=0.5\linewidth]{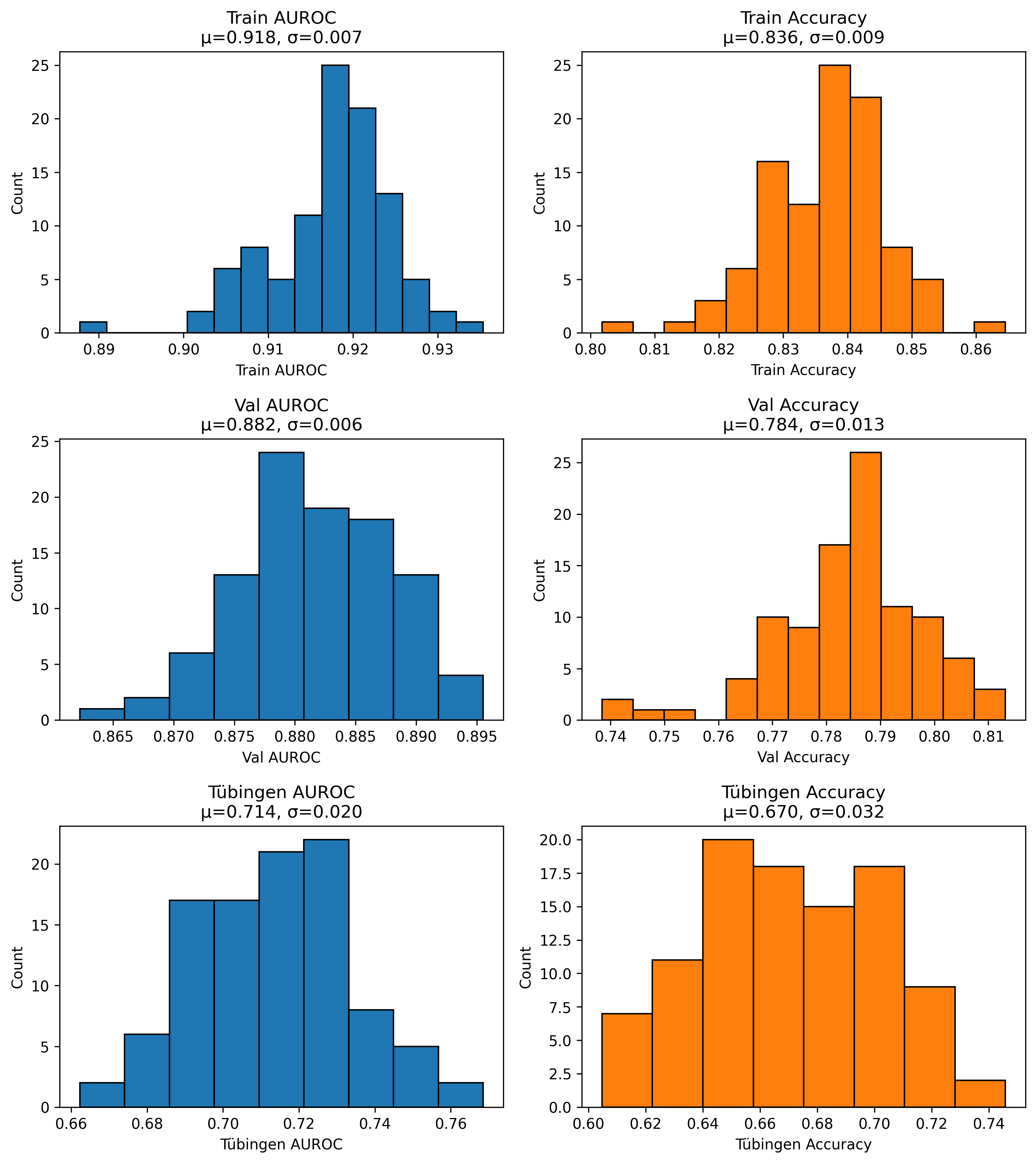}
    \caption{Distribution of results obtained by the neural network trained in the synthetic dataset (SynthNN). First row: contains the AUROC and accuracy obtained on the training set. Second row: results on the validation set (note that this is simply a different split from the synthetic dataset). Third row: distribution of results on the Tübingen dataset.}
    \label{nn_results}
\end{figure}

The key idea behind SynthNN is that if the synthetic dataset is representative of the Tübingen collection, by training a neural network on the former, it can achieve good results in the latter. However, these results displayed significant variance based on the random weight initialization performed by the neural network. Consequently, the distribution of the obtained AUROC and accuracy in the training, validation, and Tübingen sets can be seen in Figure \ref{nn_results}.

The average AUROC obtained by SynthNN in the T\"ubingen dataset is $71.4\%$, and the average accuracy is $67.0\%$. Compared with the other prominent causal discovery methods analyzed in Table \ref{methods_stats}, it can be placed squarely in the middle of the field. Methods such as SLOPE (AUROC $78.9\%$, Acc $71.5\%$) and bQCD (AUROC $73.0\%$, Acc$70.0\%$) outperform our approach, while IGCI (AUROC $70.7\%$, Acc $65.1\%$) delivers results very similar to ours. Other competitive techniques like EMD (AUROC $69.3\%$, Acc $61.7\%$) and CGNN (AUROC $66.7\%$, Acc $62.5\%$) trail our performance, indicating that our method represents a robust, although not state‑of‑the‑art, causal discovery method.

Furthermore, Singh et al. \citep{cnntueb} trained a very similar neural network directly on the Tübingen dataset, having achieved an AUROC of $76.9\%$ and an accuracy of $73.3\%$. These results represent the extent to which a neural network (with this architecture) can properly learn the causal nature present in the Tübingen data. Consequently, it should represent an upper bound for the possible performance of any similar neural network trained on synthetic data. Since the results obtained by SynthNN are not far from those obtained by training directly on the Tübingen dataset, it is reasonable to conclude that most of the remaining error should be attributed to limitations of the neural network itself. 

On the other hand, it also means the synthetic dataset should represent the dynamics of the Tübingen data accurately, at least as far as the neural network is able to capture the causal relationship between data. One can think of SynthNN as offering a lower bound on how representative the synthetic dataset is of the Tübingen data and thus of real-world causal systems.

Regardless of the point of view, it seems clear that the results obtained by SynthNN provide additional evidence of how representative the developed synthetic dataset is of real-world data. As such, this method serves as further validation, not only of the dataset itself, but also of its underlying assumption: the central role of exchangeability in causal discovery. This is an even stronger result when taking into account the many unintentional additional assumptions present in the design choices of both the synthetic dataset generation algorithm and the neural network architecture.  

Lastly, SynthNN is a causal discovery method in full right, since it estimates the causal direction of real-world data, only based on first principles. This is in stark contrast to the CNN from Singh et al. \citep{cnntueb}, which is itself trained on the Tübingen data. However, unlike other methods, such as LiNGAM or bQCD, it does not have such a direct downstream application to the problem of causal representation learning. While LiNGAM and bQCD's assumptions have concise mathematical representations and thus can be more easily fit into a machine learning pipeline, SynthNN's key assumption lies in the representational power of the synthetic dataset, which is harder to integrate. Nevertheless, SynthNN clearly shows the validity of developing a causal discovery method by training a neural network (or any other machine learning algorithm) on a dataset that is believed to represent real-world data. 

\section{Conclusion}
We have shown that the traditional i.i.d. framework for causal discovery is most naturally and more powerfully understood as exchangeable. This was first substantiated with the proposal of a new hierarchical and exchangeable causal model that simultaneously adheres to the Causal de Finetti's theorem, while enabling a deeper interpretation of the possibilities behind this approach. Specifically, the proposed model takes into consideration uncertainty regarding both the causal mechanism and the distribution of inner latent variables. Explicitly modelling these uncertainties is crucial to obtain a fairer description of the complexities of the real world. This argument is further supported by gathering inspiration from the neighbouring domain of causal inference, while an analysis of the T\"ubingen dataset heavily suggests a high number of unknown distributions of plausible latent variables.

Afterwards, to make these ideas more concrete, we introduced a novel synthetic dataset that enforces only exchangeability—eschewing stronger i.i.d. constraints—and showed that its statistical properties align more closely with the Tübingen dataset than any prior synthetic benchmark. Complementing the dataset, we presented a neural‑network–based causal‑discovery algorithm trained solely on our exchangeable examples. This model performs similarly to other relevant and current methods on real‑world data, demonstrating the viability of an exchangeability‑only approach.
Even though the broader implications of adopting exchangeability in causal‑discovery practice remain open to be fully explored, this work opens the door to novel dependencies, structures, and methodologies that more accurately reflect the complexities of real‑world data.

\section{Author's Statement}

\noindent 
\textbf{Funding Information:} This work was partially
supported by Instituto de Telecomunicações (own funds).

\noindent
\textbf{Author Contribution:} Conceptualization, TB and MB. Formal development, TB. Coding and experiments, TB. Writing original draft, TB. Revision and editing, TB and MF. Supervision, MF.

\noindent
\textbf{Conflict of Interest:} None.

\oldbibliographystyle{plainnat}
\bibliography{article_bib}

\appendix 
\newpage

\section{Tübingen Statistical Assumptions Analysis} \label{Tübingen_appendix}

This appendix provides a detailed analysis of the T\"ubingen dataset. Specifically, we analysed whether each example should be more accurately modelled using exchangeability, and the proposed hierarchical model defined in Equation \eqref{hier_exc_eq}. More so, we focused on the second condition and whether there were any relevant latent variables with unknown distributions that fit neatly into the description of exchangeable processes. Table \ref{tab:Tübingen_data} reports this analyses, alongside possible examples of these latent variables.

\begin{longtable}{p{4cm}p{4cm}p{4cm}p{2cm}}
\caption{Tübingen Dataset Analysis. The first two columns contain the cause and effect definitions (as in the source material). The latter two provide possible examples of latent variables with unknown distributions and whether the sequence should be considered exchangeable.} \label{tab:Tübingen_data} \\
\toprule
\textbf{Cause} & \textbf{Effect} & \textbf{Latent Variable Examples} & \textbf{Exchangeable?} \\
\midrule
\endhead
altitude & temperature (average over 1961-1990) & longitude, latitude & yes \\
altitude & precipitation (yearly value averaged over 1961-1990) & longitude, latitude & yes \\
longitude & temperature (averaged over 1961-1990) & altitude, latitude & yes \\
altitude & sunshine (yearly value averaged over 1961-1990) & longitude, latitude & yes \\
Oyster age (estimated using its rings) & Longest shell measurement & genes & yes \\
Oyster age (estimated using its rings) & Shell weight & genes & yes \\
Oyster age (estimated using its rings) & Diameter & genes & yes \\
Oyster age (estimated using its rings) & Height & genes & yes \\
Oyster age (estimated using its rings) & Whole weight & genes & yes \\
Oyster age (estimated using its rings) & Shucked weight & genes & yes \\
Oyster age (estimated using its rings) & Viscera weight & genes & yes \\
Age & Wage per hour & education & yes \\
displacement & mpg & horsepower, weight & yes \\
horsepower & mpg & weight, displacement & yes \\
weight & mpg & horsepower, displacement & yes \\
horsepower & acceleration & weight, displacement & yes \\
Age & Dividends from stock & education & yes \\
age of child in years & concentration of GAG & genes & yes \\
duration of erruption in minutes & time to the next erruption in minutes & weather & yes \\
latitude & temperature (averaged over 1961-1990) & altitude, longitude & yes \\
longitude & precipitation (yearly value averaged over 1961-1990) & altitude, latitude & yes \\
age & height & genes & yes \\
age & weight & genes & yes \\
age & heart rate & genes & yes \\
cement & compressive strength & other mixture components & yes \\
blast furnace slag & compressive strength & other mixture components & yes \\
fly ash & compressive strength & other mixture components & yes \\
water & compressive strength & other mixture components & yes \\
superplasticizer & compressive strength & other mixture components & yes \\
coarse aggregate & compressive strength & other mixture components & yes \\
fine aggregate & compressive strength & other mixture components & yes \\
age & compressive strength & other mixture components & yes \\
alcoholic comsumption & mean corpuscular volume & genes & yes \\
alcoholic comsumption & alkphos & genes & yes \\
alcoholic comsumption & sgpt & genes & yes \\
alcoholic comsumption & sgot & genes & yes \\
alcoholic comsumption & gammagt & genes & yes \\
age & body mass index (weight in kg/(height in m)$^2$) & genes & yes \\
age & 2-Hour serum insulin (mu U/ml) & genes & yes \\
age & diastolic blood pressure (mm Hg) & genes & yes \\
age & Plasma glucose concentration a 2 hours in an oral glucose tolerance test & genes & yes \\
days of the year & mean daily temperature of Furtwangen & location & time series \\
temperature: year 2000, day 50 & temperature: year 2000, day 51 & location & yes \\
temperature: year 2000, day 50 & temperature: year 2000, day 51 & location & yes \\
temperature: year 2000, day 50 & temperature: year 2000, day 51 & location & yes \\
temperature: year 2000, day 50 & temperature: year 2000, day 51 & location & yes \\
weekend? (binary data) & number of cars per 24h at different counting stations in Oberschwaben, Germany & weather & yes \\
Outdoor temperature & Indoor temperature & time series & yes \\
Temperature (degree celsius) & Ozone (microgram / cubic meter) & Car gases emission & yes \\
Temperature (degree celsius) & Ozone (microgram / cubic meter) & Car gases emission & yes \\
Temperature (degrees Celsius), Davos-See, Switzerland & Ozone (microgram / cubic meter), Davos-See, Switzerland & Car gases emission & yes \\
day 50: 4 metereological variables & day 51: 4 metereological variables & location & yes \\
(wind speed ($m/s$), global radiation ($W/m^2$), temperature) & Ozon concentration ($microgramm/m^3$) & Car gases emission & yes \\
(displacement, horsepower, weight) & (mpg, acceleration) & aerodynamics & yes \\
Temperature (degree celsius) & Ozone (microgram / cubic meter) & Car gases emission & yes \\
latitude of the country's capital & life expectancy at birth for different countries, female, 2000-2005 & longitude & yes \\
latitude of the country's capital & life expectancy at birth for different countries, female, 1995-2000 & longitude & yes \\
latitude of the country's capital & life expectancy at birth for different countries, female, 1990-1995 & longitude & yes \\
latitude of the country's capital & life expectancy at birth for different countries, female, 1985-1990 & longitude & yes \\
latitude of the country's capital & life expectancy at birth for different countries, male, 2000-2005 & longitude & yes \\
latitude of the country's capital & life expectancy at birth for different countries, male, 1995-2000 & longitude & yes \\
latitude of the country's capital & life expectancy at birth for different countries, male, 1990-1995 & longitude & yes \\
latitude of the country's capital & life expectancy at birth for different countries, male, 1985-1990 & longitude & yes \\
Population with sustainable access to improved drinking water sources (\%) total, 2006 & Infant mortality rate (per 1 000 live births) both sexes, 2006 & health car access & yes \\
stock returns of Hang Seng Bank (0011.HK) & stock return of HSBC Hldgs (0005.HK) & other HSBC holdings & yes \\
stock returns of Hutchison (0013.HK) & stock return of Cheung kong (0001.HK) & other Cheung kong holdings & yes \\
stock returns of Cheung Kong (0001.HK) & stock return of Sun Hung Kai Prop. (0016.HK) & other Sun Hung Kai Prop. holdings & yes \\
open http connections during that minute & bytes sent at minute t & user type & yes \\
outside temperature in degrees Celsius & inside room temperature in degrees Celsius &  & time series \\
par (between 0 and 14, 0 $->$ very female, 14 $->$ very male) & sex guess (0: female or 1: male, the subject's guess) & design choices & yes \\
(Temperature of patient { 35C-42C }, Occurrence of nausea, Lumbar pain, Urine pushing (continuous need for urination), Micturition pains, Burning of urethra, itch, swelling of urethra outlet) & (decision: Inflammation of urinary bladder, decision: Nephritis of renal pelvis origin) & age, doctor & yes \\
sunspot area & global mean temperature anomalies (deviations from 1961-1990) in °C & location & yes \\
Energy use (kg of oil equivalent per capita) for different countries in different years & CO2 emissions for different countries in different years & country development level & yes \\
GNI (Gross national income) per capita for different countries (in US\$) & life expectancy at birth for different countries & health car access & yes \\
GNI (Gross national income) per capita for different countries (in US\$) & under 5 mortality rate for different countries (deaths per 1000 live births) & health car access & yes \\
the average annual rate of change of population & the average annual rate of change of total dietary consumption for total population (kcal/day) & rate of change of individual dietary consumption & yes \\
the solar radiation in W/$m^2$ & the daily average temperature of the air measured at the same location and the same days & latitude & time series \\
PPFD (Photosynthetic Photon Flux Density) & NEP (Net Ecosystem Productivity) & Gross Primary Productivity (GPP) & yes \\
PPFDdif (Photosynthetic Photon Flux Density, diffusive) & NEP (Net Ecosystem Productivity) & PPFDdir (Photosynthetic Photon Flux Density, direct) & yes \\
PPFDdir (Photosynthetic Photon Flux Density, direct) & NEP (Net Ecosystem Productivity) & PPFDdif (Photosynthetic Photon Flux Density, diffusive) & yes \\
Temperature in degree Celsius & CO2 flux at night &  & time series \\
Temperature in degree Celsius & CO2 flux at night & & time series \\
Temperature in degree Celsius & CO2 flux at night &  & time series \\
the natural logarithm of the corresponding population & the natural logarithm of employment in 1980 in 3102 counties in US & wealth & yes \\
time to take weekly measurements (from 1 to 14) & protein content of the milk produced by each cow at time X & genes & yes \\
size in m$^2$ of appartment/room & monthly rent in EUR & location & yes \\
MeanTemp (deg Celsius) & TotalSnow (cm) & precipitation & yes \\
age & Relative Spinal bone mineral density & genes & yes \\
Mass loss OCTOBER 2012 in \% & Mass loss APRIL 2012 in \% & ecosystem & yes \\
Mass loss OCTOBER 2012 in \% & Mass loss APRIL 2012 in \% & ecosystem & yes \\
Clay content in soil (in gram per g/kg) & Soil moisture at 10cm depth (in \%) & precipitation & yes \\
Organic C content in soil (in g Carbon/kg) & Clay content (in g/kg) & concentration of other atoms & yes \\
average precipitation over 1948 to 2004 in mm/day & average runoff in over 1948 to 2004 mm/day & drainage system quality & yes \\
hour of the day & temperature in degree celsius & location & time series \\
hour of the day & load: the total electricity consumption in a region of Turkey in "MWh" & location & time series \\
temperature in degree celsius & load: the total electricity consumption in a region of Turkey in "MWh" & location & time series \\
initial speed of a ball on a ball track for children & final speed of a ball on a ball track for children & none & no \\
initial speed of a ball on a ball track for children & final speed of a ball on a ball track for children & none & no \\
social-economic status of pupil's family & language test score & IQ & yes \\
cycle time in nanoseconds & published performance on a benchmark mix relative to an IBM 370/158-3 & energy consumption & yes \\
grey value of a pixel that is chosen randomly from a fixed image. The grey value & light intensity seen by a photo diode placed several centimeters away from the screen. & none & no \\
position on the ball track where the ball starts & time interval between passing the first and the second light barrier & none & no \\
position on the ball track where the ball starts & time interval between passing the third and the fourth light barrier & none & no \\
time interval between passing the third and the fourth light barrier & time interval between passing the third and the fourth light barrier & none & no \\
pixel vector of grey values of the patch & light intensity seen by a photo diode placed several centimeters away from the screen & none & no \\
electric voltage & time required for passing one round & none & no \\
contrast & answer correct or not & observer & yes \\
time for 1/6 rotation & temperature in Degree Celsius & none & no \\
\bottomrule
\end{longtable}

\newpage

\section{Synthetic Dataset Generation}

In this appendix, we first provide details on the 8 implemented functional mechanisms, controlled by the exchangeable latent variable $\psi_2$. Afterwards, we provide 32 random examples of causal pairs generated by the presented synthetic process..

\subsection{Considered Functional Causal Mechanisms} \label{algorithm_extras}

In order to describe the 8 families of mathematical functions considered to represent the causal mechanism, we must first discuss the representation of their parameters. These are also defined by $\psi_2$, and they take different forms, depending on the specific family of functions under consideration. The controllable hyperparameters controlling the distributions from which these specific function parameters are sampled will be called $\lambda$. These were tuned to ensure that different parameter samples resulted in significant changes to the shape of the function, while ensuring both $X_n$ and $\gamma_n$ made significant contributions to the value of $Y_n$. This introduced randomness results in a more diverse and robust dataset.

For four of the implemented functions, the parameters will define the mean and standard deviation of $\gamma$. Remember that $\gamma$ had been min-max scaled to the interval $[0,1]$. After sampling the function parameters, $\gamma$ is rescaled using $\{\mu_{\gamma},\sigma_{\gamma}\}$, to ensure it has the desired mean and standard deviation. This rescaled distribution will be called $a$. Consequently, the hyperparameters define the distributions (and respective parameters) of both $\{\mu_{\gamma},\sigma_{\gamma}\}$. In all, this can be written as:
\begin{align}
    \lambda&=\{F_\mu, F_\sigma, \mu_\mu, \sigma_\mu, \mu_\sigma, \sigma_\sigma \}, \\
    \mu_{\psi_2} \sim &F_\mu (\mu_\mu, \sigma_\mu) ; \,\, \sigma_\gamma \sim F_{\psi_2} (\mu_\sigma, \sigma_\sigma), \\
    a&=\mbox{Rescale}(\gamma;\mu_{\psi_2},\sigma_{\psi_2}),
\end{align}
where $F_\mu, F_\sigma$ define the probability distributions (either Gaussian or Uniform), and $ \mu_\mu, \sigma_\mu, \mu_\sigma, \sigma_\sigma$ their parameters.

In specific, given $a$, these four functions are defined as: 
\begin{enumerate}

  \item \textbf{Exponential} ($f_{\mbox{exp}}$):
  \begin{equation}
      Y_n=e^{X_na_n}.
  \end{equation}

  \item \textbf{Logarithmic} ($f_{\log}$):
 \begin{equation}
      y_n=\log(x_n+a_n).
  \end{equation}
  
  \item \textbf{Inversely proportional} ($f_{\mbox{inv}}$): 
  \begin{equation}
      y_n=\frac{1}{x_n+a_n}.
  \end{equation}

  \item \textbf{Power law} ($f_{\mathrm{pow}}$): 

  \begin{equation}
      y_n=x_n^{a_n}.
  \end{equation}
\end{enumerate}

For these mathematical families, the function parameters serve only to alter $\gamma$'s distribution. So, it would be fair to wonder why not include these parameters directly in $\gamma$'s distribution, defined by $\psi_1$. The current approach was preferred because 1) it allows for more flexibility, since the distribution of $\gamma$ can depend on $X_n$; and 2) it clears up the logic, leaving all detailed function parameters to $\psi_2$.

Having said this, for three more families of functions, the parameters will simultaneously control the rescaling process of $\gamma$ (as is done for the first four), and specific function criteria. The latter will be discussed for each function in time. Regarding the first, these four models will implement, even if in different ways, the same linear approach to rescale $\gamma$. The core idea is to sample (in accordance with 
$\lambda$) the mean and standard deviations ($\mu_\text{start}, \sigma_\text{start},\mu_\text{end}$ and $\sigma_\text{end}$), associated with two (previously chosen) values of $X$ (called $X_\text{start}$ and $X_\text{end}$). The mean and standard deviation for $\gamma$ will depend on $X_n$, and it will be given by the weighted average of the mean and standard deviations at $X_\text{start}$ and $X_\text{end}$, in proportion to their distance. Mathematically:
\begin{align}
    a_\text{start}=\mbox{Rescale}(\gamma;\mu_\text{start},\sigma_\text{start}), \,\, a_\text{end}=\mbox{Rescale}(\gamma;\mu_\text{end}&,\sigma_\text{end}),\\
    a_n= \frac{a_{\text{end}_n} (X_\text{end}-X_n)+  a_{\text{start}_n}(X_n-X_\text{start})}{X_\text{end}-X_\text{start}},
\end{align}
Additionally, these three families of functions will always consider a linear relationship between $a_n$ (the rescaled $\gamma_n$) and $X_n$,
\begin{equation}
    Y_n=a_nX_n.
\end{equation}

For the following three mathematical functions, the specific parameters that must be defined are: 1) which values will $X_\text{start}$ and $X_\text{end}$ take, and 2) how are $\mu_\text{start}, \sigma_\text{start},\mu_\text{end}$ and $\sigma_\text{end}$ sampled. Additionally, some families of functions will break up the real axis into multiple intervals, each with their own $X_\text{start}$ and $X_\text{end}$. Since each boundary will necessarily serve as the right boundary of an interval and the left of the next one, we will use the notation $X_k$ to represent these points.  Having said this, the three families of functions under consideration are:
\begin{enumerate}
\setcounter{enumi}{4}
   \item \textbf{Linear} ($f_{\mbox{lin}}$):
   \begin{align}
        X_\text{start}=0, \,X_\text{end}=1, \\
       \sigma_\text{start},\sigma_\text{end}\sim \text{Inv-Gamma}\bigl(\nu_{m},\,\nu_{v}\bigr),\\
    \mu_\text{start} = 0 \quad  \mu_\text{end} = \tan\phi \quad  \phi \sim \mathcal{U}(0,2\pi),
   \end{align}
   where $\mbox{Inv-Gamma}\bigl(\nu_{m},\,\nu_{v}\bigr)$ stands for the Inverse Gamma distribution with fixed hyperparameters $\lambda=\{\nu_{m},\,\nu_{v}\}$, and $\mathcal{U}$ is the uniform distribution.

  \item \textbf{Piecewise linear} ($f_{\mbox{mix}}$):  
  \begin{align}
      K \sim \mathcal{U}(2,\mbox{max}_K), \\
      \{X_k\} = \{X_l \sim \mathcal{U}(0,1)\}_K \cup \{0,1\}, \\ 
      \sigma_k\sim \text{Inv-Gamma}\bigl(\nu_{m},\,\nu_{v}\bigr), \,\, \phi \sim \mathcal{U}(0,2\pi), \\
      \mu_0=0, \mu_{k+1}=\mu_k+\tan(\phi)(X_{k+1}-X_k),
  \end{align}
    where $\lambda=\{\mbox{max}_K,\nu_{m},\,\nu_{v}\}$ are the function's hyperparameters.
    
  \item \textbf{Brownian‐like motion} ($f_{\mbox{brown}}$):
  \begin{align}
      \{X_k\} =X, \\
      \mu_0=0, \quad \mu_{k+1}\sim \mathcal{N} \left(\mu_k + \frac{d\mu_k}{dt}\Delta_t, \sqrt{\frac{\Delta_t^3}{3}}\right),\\
      \hat{\sigma}_0 \sim \text{Inv-Gamma}\bigl(\nu_{m},\,\nu_{v}\bigr),\quad \sigma_0= \hat{\sigma}_0,\,\, \\\hat{\sigma}_{k+1} =
      \hat{\sigma}_k + \frac{d\hat{\sigma}_k}{dt}\Delta_t , \quad \sigma_{k+1} 
      \sim \mathcal{N}\left(\hat{\sigma}_{k+1}, \sqrt{\frac{\Delta_t^3}{3}}\right)
  \end{align}
  where $\lambda=\{\nu_{m},\,\nu_{v}\}$ are the function specific hyperparameters, and $\frac{d\mu_k}{dt}, \frac{d\hat{\sigma}_k}{dt}$ are computed using a polynomial approximation.
\end{enumerate}

Finally, the approach for the polynomial function shares many similarities with the previous three. It also starts by obtaining the rescaled $\gamma$ for different values of $X$ (called $a_k$). Then, for each $n$, it will fit a polynomial based on the pairs $\{X_k,a_{k_n}\}$. Then, it will use the obtained coefficients to obtain $Y_n$. Mathematically, 
\begin{enumerate}
\setcounter{enumi}{7}
  \item \textbf{Polynomial} ($f_{\mbox{poly}}$):
  \begin{align}
       K \sim \mathcal{U}(2,\mbox{max}_K), \\
      \{X_k\} = \{X_l \sim \mathcal{U}(0,1)\}_K \cup \{0,1\}, \\ 
      \sigma_k\sim \text{Inv-Gamma}\bigl(\nu_{m},\,\nu_{v}\bigr), \,\, \phi \sim \mathcal{U}(0,2\pi), \\
      \mu_0=0, \mu_{k+1}=\mu_k+\tan(\phi)(X_{k+1}-X_k),\\
      \{p_n\} = \text{fit}(\{X_k,a_{k_n}\}),\\
      Y_n=p_n(0)+p_n(1)X_n + p_n(2) X_n^2 + ... p_n(K)X_n^K.
  \end{align}
  where $\lambda=\{\nu_{m},\,\nu_{v}\}$ are the function specific hyperparameters, and the fit function solves a determined system.
\end{enumerate}

Additionally, in functions whose design is monotone, a Bernoulli fair coin is tossed once per example to decide whether to flip the sign of $Y$ (and, for some mechanisms, of~$X$ as well) so that both monotone orientations occur with equal frequency. 

\subsection{Examples from the Synthetic Dataset} \label{examples_appendix}

\begin{figure}[H] 
    \centering
    \includegraphics[width=0.95\linewidth]{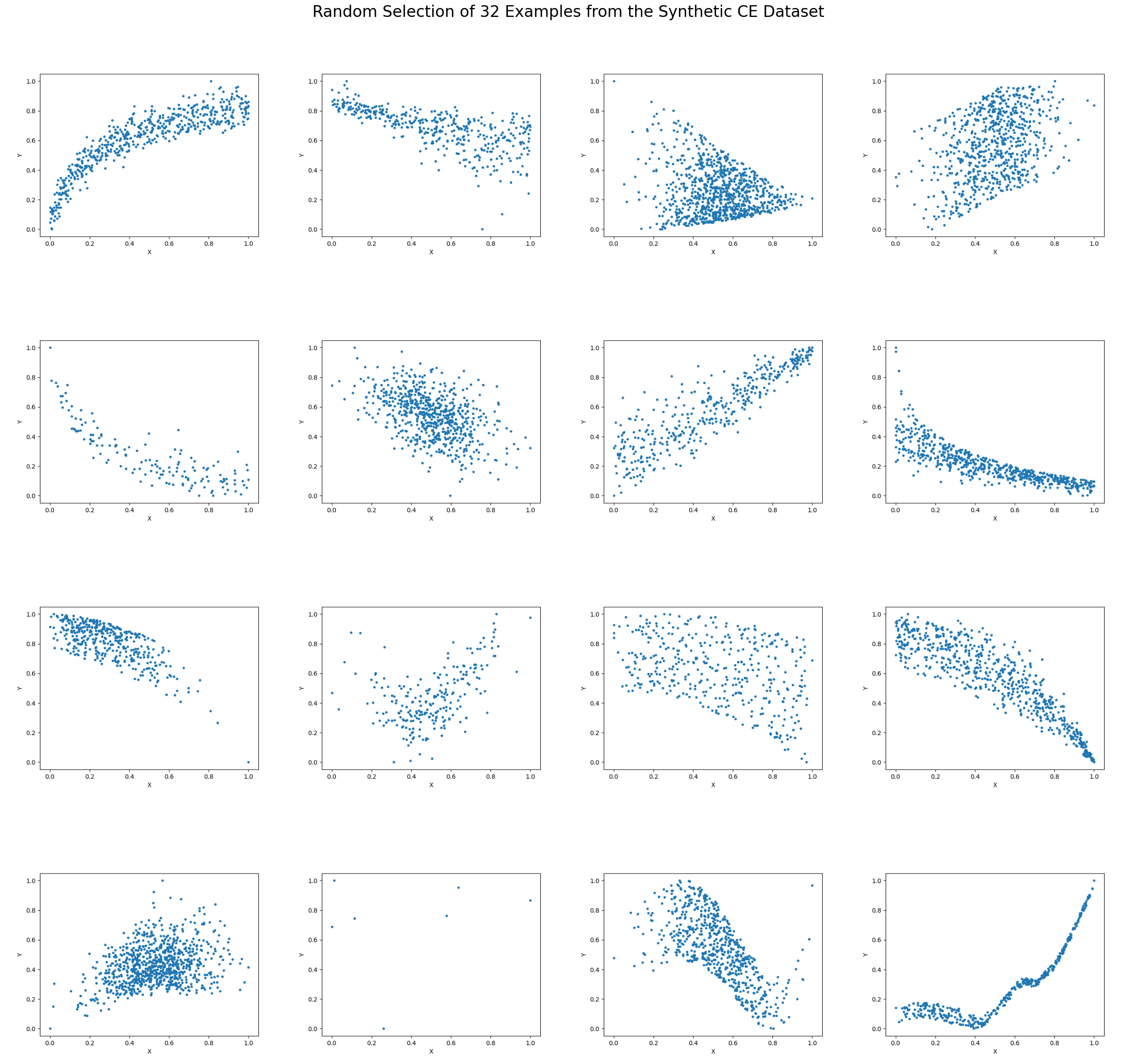}
    \caption{32 examples randomly sampled from the developed synthetic dataset.}
\end{figure}

\newpage

\section{SynthNN specifications}\label{nn_specs}

In this appendix, we provide the full technical description of the convolutional neural network (CNN) used for binary classification, including data preprocessing, labeling procedures, and relevant implementation details omitted from the main text.

Additionally, in bivariate causal discovery datasets, even though the cause and effect are well-defined, there is an inherent arbitrariness in the labeling process (which variable should be considered $X$, which $Y$, and consequently, the true label $X \to Y$ or $Y \to X$). Therefore, to ensure an unbiased training of the developed neural network, we must ensure a balanced weighted training dataset (with $50\%$ $X\xrightarrow{}Y$ and $50\%$ $Y\xrightarrow{}X$).

\subsection{Data preprocessing}

The data is originally presented as two paired vectors ($X$ and $Y$). This is converted into a binary 50x50 image ($N=50$) using the following steps:
\begin{enumerate}
  \item Min–\mbox{max} scale both $X$ and $Y$ to \([0,1]\).
  \item Convert scaled coordinates to integer pixel indices:
  \[
    i_n = \lfloor X_n\,(N-1)\rfloor,\quad 
    j_n = \lfloor Y_n\,(N-1)\rfloor.
  \]
  \item Initialize an \(N\times N\) zero matrix and set \(I[j,i] = 1\) for each obtained point \((i_n,j_n)\).
\end{enumerate}

Finally, during training, a Gaussian filter is applied to the images with $\sigma=0.5$.

\subsection{Architecture}

The neural network has a total of 1,739,777 trainable parameters, containing:
\begin{itemize}
  \item \textbf{Convolutional blocks:} three 2D convolutional layers (3×3 kernels), each followed by a 2×2 max‐pooling, with filter counts doubling each block (32, 64, 128).
  \item \textbf{Dense layers:} three fully connected layers of sizes 256, 128, and 64, respectively.
  \item \textbf{Activation:} ReLU for all hidden layers, sigmoid activation for the final output neuron.
  \item \textbf{Regularization:} $\ell_2$ weight decay with coefficient $\lambda = 0.01$ on all kernels.
  \item \textbf{Loss function:} binary cross‐entropy.
  \item \textbf{Optimizer:} Adam \(\bigl(\alpha = 10^{-4}\bigr)\).
\end{itemize}

The sequential architecture of SynthNN can be seen in Table \ref{nn_layers}.
\begin{table}[h!]
  \centering
  \begin{tabular}{lccc}
    \toprule
    \textbf{Layer (type)} & \textbf{Output Shape} & \textbf{\# Parameters} & \\
    \midrule
    Conv2D (3×3, 32 filters)           & (None, 50, 50, 32)     &    320  \\
    MaxPooling2D (2×2)                 & (None, 25, 25, 32)     &      0  \\
    Conv2D (3×3, 64 filters)           & (None, 25, 25, 64)     & 18,496  \\
    MaxPooling2D (2×2)                 & (None, 13, 13, 64)     &      0  \\
    Conv2D (3×3, 128 filters)          & (None, 13, 13, 128)    & 73,856  \\
    MaxPooling2D (2×2)                 & (None, 7, 7, 128)      &      0  \\
    Flatten                            & (None, 6\,272)         &      0  \\
    Dense (256 units)                  & (None, 256)            & 1,605,888 \\
    Dense (128 units)                  & (None, 128)            &  32,896 \\
    Dense (64 units)                   & (None, 64)             &   8,256 \\
    Dense (1 unit, sigmoid)            & (None, 1)              &      65 \\
    \bottomrule
  \end{tabular}
  \caption{Detailed layer-by-layer summary of SynthNN}
  \label{nn_layers}
\end{table}

\subsection{Evaluation on the Tübingen dataset}

The output of the neural network represents the estimated posterior probability that the input data has the causal direction $X\xrightarrow{}Y$. However, in order to improve prediction consistency at test time, the following procedure is implemented:

\begin{enumerate}
  \item Remove outliers beyond the $90\%$ quantile.
  \item Generate two images per pair: one for \((X,Y)\) and one for \((Y,X)\).
  \item Obtain model predictions \(\hat p_{X\xrightarrow{}Y}\), \(\hat p_{Y\xrightarrow{}X}\).
  \item Output the asymmetry score:
    \[
      s = \frac{\hat p_{X\xrightarrow{}Y} - \hat p_{Y\xrightarrow{}X}}{\hat p_{X\xrightarrow{}Y} + \hat p_{Y\xrightarrow{}X}}.
    \]
\end{enumerate}

\end{document}